\documentclass[10pt,twocolumn,letterpaper]{article}

\usepackage{cvpr}
\usepackage{times}
\usepackage{epsfig}
\usepackage{graphicx}
\usepackage{amsmath}
\usepackage{amssymb}
\usepackage{array}

\usepackage{float}

\usepackage{wrapfig}
\usepackage{stackengine}
\usepackage{xcolor}
\usepackage{rotating}
\usepackage[export]{adjustbox}
\usepackage{xspace}
\makeatletter
\usepackage{caption}
\usepackage{booktabs}       
\usepackage{tabu}
\usepackage{multirow}
\usepackage{arydshln}
\usepackage{authblk}
\usepackage{gensymb}

\newcommand{\mycaption}[2]{\caption{\textbf{#1.}\xspace#2}}

\newcommand{\authornote}[3]{}

\newcommand{\paragrapht}[1]{\noindent {\textbf{#1}}}
\def\ie{i.e.,\ }
\def\eg{e.g.,\ }

\usepackage[pagebackref=false,breaklinks=true,letterpaper=true,colorlinks,urlcolor=blue,citecolor=blue,linkcolor=blue,bookmarks=false]{hyperref}

\cvprfinalcopy 

\def\cvprPaperID{68} 

\ifcvprfinal\fi
\begin{document}

\title{SCOPS: Self-Supervised Co-Part Segmentation}
\author[1*]{Wei-Chih Hung}
\author[2]{Varun Jampani}
\author[2]{Sifei Liu}
\author[2]{Pavlo Molchanov}
\author[1]{Ming-Hsuan Yang}
\author[2]{Jan Kautz}
\makeatletter 
\renewcommand\AB@affilsepx{\quad \protect\Affilfont} 
\makeatother
\affil[1]{UC Merced}
\affil[2]{NVIDIA}


\newcommand\blfootnote[1]{%
  \begingroup
  \renewcommand\thefootnote{}\footnote{#1}%
  \addtocounter{footnote}{-1}%
  \endgroup
}

\twocolumn[{%
\renewcommand\twocolumn[1][]{#1}%
\maketitle
\begin{center}
\vspace{-6mm}
  
\newlength{\teaserw}
\setlength{\teaserw}{0.105\linewidth}
\newlength{\teaserh}
\setlength{\teaserh}{0.12\linewidth}

\begin{tabular}
{   @{\hspace{0mm}}c@{\hspace{0.3mm}} 
    @{\hspace{0mm}}c@{\hspace{0.3mm}}
    @{\hspace{0mm}}c@{\hspace{2.0mm}}
    @{\hspace{0mm}}c@{\hspace{0.3mm}} 
    @{\hspace{0mm}}c@{\hspace{0.3mm}}
    @{\hspace{0mm}}c@{\hspace{2.0mm}}
    @{\hspace{0mm}}c@{\hspace{0.3mm}}
    @{\hspace{0mm}}c@{\hspace{0.3mm}}
    @{\hspace{0mm}}c@{\hspace{0.3mm}} 
}
    \includegraphics[width=\teaserw,height=\teaserh]{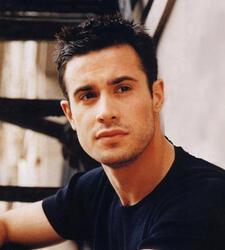} &
    \includegraphics[width=\teaserw,height=\teaserh]{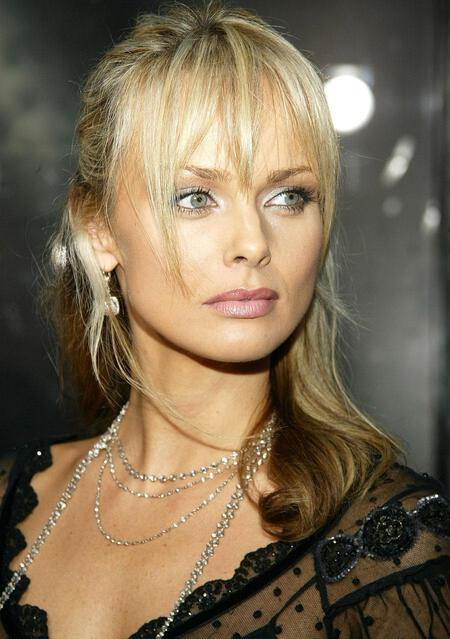} &
    \includegraphics[width=\teaserw,height=\teaserh]{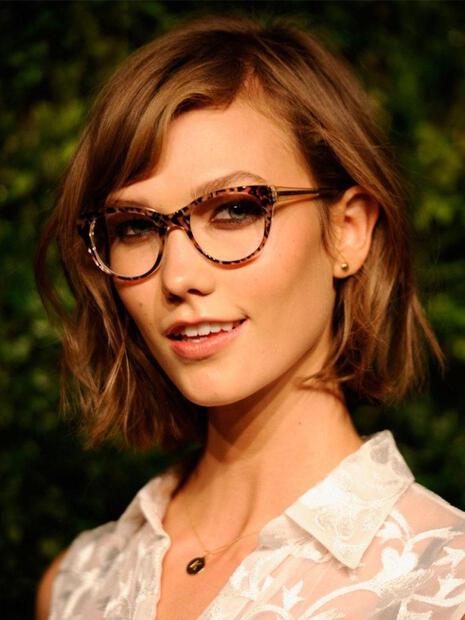} &
    
    \includegraphics[width=\teaserw,height=\teaserh]{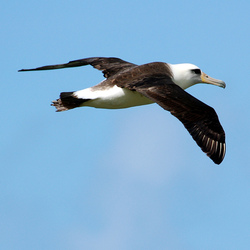} &
    \includegraphics[width=\teaserw,height=\teaserh]{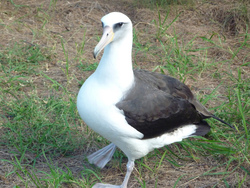} &
    \includegraphics[width=\teaserw,height=\teaserh]{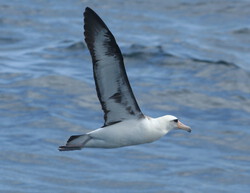} &
    
    \includegraphics[width=\teaserw,height=\teaserh]{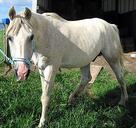} &
    \includegraphics[width=\teaserw,height=\teaserh]{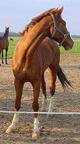} &
    \includegraphics[width=\teaserw,height=\teaserh]{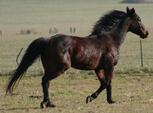} \\

    \includegraphics[width=\teaserw,height=\teaserh]{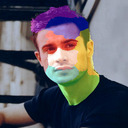} &
    \includegraphics[width=\teaserw,height=\teaserh]{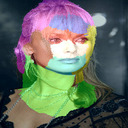} &
    \includegraphics[width=\teaserw,height=\teaserh]{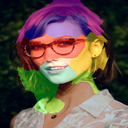} &

    \includegraphics[width=\teaserw,height=\teaserh]{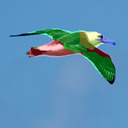} &
    \includegraphics[width=\teaserw,height=\teaserh]{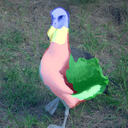} &
    \includegraphics[width=\teaserw,height=\teaserh]{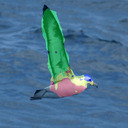} &
    
    \includegraphics[width=\teaserw,height=\teaserh]{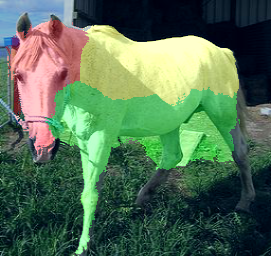} &
    \includegraphics[width=\teaserw,height=\teaserh]{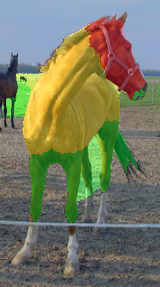} &
    \includegraphics[width=\teaserw,height=\teaserh]{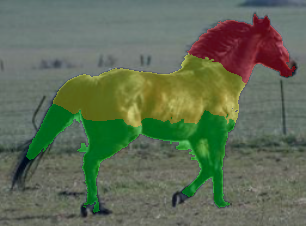} \\

\end{tabular}
\captionof{figure}{\textbf{Robustness to variations.} Sample part segmentation obtained by SCOPS on different types of image collections: (left) unaligned faces from CelebA~\cite{liu2015faceattributes}, (middle) birds from CUB~\cite{wah2011caltech} and (right) horses from PASCAL  VOC~\cite{PASCAL_VOC_2010} dataset images, showing that SCOPS can be robust to appearance, viewpoint and pose variations.}
\label{fig:teaser}
\vspace{8mm}

  \vspace{-6mm}
\end{center}%
}]

\begin{abstract}
Parts provide a good intermediate representation of objects 
that is robust with respect to the camera, pose and appearance variations.
Existing works on part segmentation is dominated by supervised
approaches that rely on large amounts of manual annotations and 
can not generalize to unseen object categories. We propose
a self-supervised deep learning approach for part segmentation,
where we devise several loss functions that aids in predicting
part segments that are geometrically concentrated, robust to object 
variations and are also semantically consistent across different
object instances. Extensive experiments on different types of 
image collections demonstrate that our approach can produce
part segments that adhere to object boundaries and also more
semantically consistent across object instances compared to
existing self-supervised techniques.
\end{abstract}

\blfootnote{*This work is done when the author was doing internship at NVIDIA.}

\vspace{-10mm}




\vspace{-3mm}
\section{Introduction}
\label{sec:intro}
\vspace{-1mm}

Much of the computer vision involves analyzing objects surrounding
us, such as humans, cars, furniture, and so on. A major challenge in
analyzing objects is to develop a model that is robust to the
multitude of object transformations and deformations due to changes
in camera pose, occlusions, object appearance, and pose variations.
Parts provide a good intermediate representation of objects that
is robust with respect to these variations. As a result, part-based representations
are used in a wide range of object analysis tasks such as
3D reconstruction~\cite{zuffi2015stitched}, detection~\cite{felzenszwalb2008discriminatively},
fine-grained recognition~\cite{krause20133d}, pose estimation~\cite{kiefel2014human}, etc.

Several types of 2D part representations have been used in the
literature, with the three most common ones being landmarks,
bounding boxes, and part segmentations. A common approach to
the part analysis is to first manually annotate large amounts of data
and then leverage fully-supervised approaches to recognize parts~\cite{chen_cvpr14,liu2015faceattributes, azizpour2012object, bourdev2009poselets, branson2011strong}.
However, these annotations, especially part segmentation,
are often quite costly.
The annotations are also specific to a single object category
and usually do not generalize to other object classes.
Consequently, it is difficult to scale the fully-supervised models
to unseen categories and there is a need for weakly supervised
techniques for part recognition that only rely on very weak
supervision or no supervision at all.

Part representations, once obtained, are robust to
variations and help in high-level object understanding.
However, obtaining part segmentations is challenging
due to the above mentioned intra-class variations.
An image collection of a single object category, despite having
the same category objects, usually have high variability regarding pose, object appearances, camera viewpoint, the presence of multiple objects, etc. Figure~\ref{fig:teaser} shows some sample images from three different image collections. Notice the variability across different object instances. Any weakly or
unsupervised technique for part segmentation needs to reason about correspondences between different images which is challenging in such diverse image collections.

In this work, we propose a self-supervised deep learning framework for
part segmentation. Given only an image collection of the same object category, our model can learn part segmentations that are semantically consistent across different object instances. Our learning technique is class agnostic, \ie can be applied to any type of rigid or non-rigid object categories. And, we only use very weak supervision in the form of ImageNet pre-trained features~\cite{Krizhevsky_NIPS_2012,vgg,he2016deep}, which are readily available.
Contrary to recent deep learning techniques~\cite{thewlis2017unsupervised,thewlis2017unsupervised2,zhang2018unsupervised}, which learn landmarks (keypoints) in a weakly or un-supervised manner, our network predicts part segmentation which provides much richer intermediate object representation compared to landmarks or bounding boxes.

To train our segmentation network, we consider several properties
of a good part segmentation and encode that prior knowledge
into the loss functions. Specifically, we consider four
desirable characteristics of a part segmentation:

\vspace{-1mm}
\begin{itemize}
    \vspace{-2mm}
    \item \emph{Geometric concentration}: Parts are 
    concentrated geometrically and form connected components.
    \vspace{-2mm}
    \item \emph{Robustness to variations}: Part segments are robust with respect to
    object deformations due to pose changes as well as camera and 
    viewpoint changes.
    \vspace{-2mm}
    \item  \emph{Semantic consistency}: Part segments should be
    semantically consistent across different object instances 
    with appearance and pose variations.
    \vspace{-2mm}
    \item \emph{Objects as union of parts}: Parts appear on
    objects (not background) and the union of parts forms an object.
\end{itemize}
\vspace{-2mm}

We devise loss functions that favor part segmentations that has above-mentioned qualities and use these loss functions to train our part segmentation network. We discuss these loss functions in detail in Section~\ref{sec:methods}.
We call our part segmentation network ``SCOPS" (Self-Supervised Co-Part Segmentation).
Figure~\ref{fig:teaser} shows sample image collections and the corresponding 
part segmentations that SCOPS predicts. 
These visual results indicate that
SCOPS can estimate part segmentations that are semantically
consistent across object instances despite large variability
across object instances.

When compared to recent unsupervised landmark detection approaches~\cite{thewlis2017unsupervised,thewlis2017unsupervised2,zhang2018unsupervised},
our approach is relatively robust to
appearance variations while also handling occlusions. Moreover,
our approach can handle multiple object instances in an image
which is not possible via landmark estimation with a fixed number of landmarks.
When compared to the recent Deep Feature Factorization (DFF) approach~\cite{collins2018deep}, ours can scale to larger datasets, 
can produce sharper part segments that adhere to
object boundaries and also more semantically consistent across
object instances. We quantitatively evaluate our part segmentation
results with an indirect measure of landmark estimation accuracy
on unaligned CelebA~\cite{liu2015faceattributes}, AFLW~\cite{koestinger2011annotated} and CUB~\cite{wah2011caltech} dataset images,
and also with foreground segmentation accuracy on the PASCAL VOC dataset~\cite{PASCAL_VOC_2010}.
Results indicate that SCOPS consistently performs favorably against
recent techniques.
In summary, we propose a self-supervised deep network
that can predict part segmentations that are semantically consistent
across object instances while being relatively robust to object
pose and appearance variations, camera variations and occlusions.








\vspace{-3mm}
\section{Related Works}
\label{sec:related}

\vspace{-1mm}
\noindent \textbf{Object concept discovery}
CNNs have shown impressive generalization capabilities across different
computer vision tasks~\cite{yosinski2014transferable,sharif2014cnn,agrawal2014analyzing}.
As a result, several works try to interpret and visualize the
intermediate CNN representations
~\cite{zeiler2014visualizing,zhou2014object,bau2017network}.
While some recent works~\cite{gonzalez2018semantic,bau2017network} demonstrate the presence of object part information in pre-trained
CNN features, we aim to train a CNN that can predict consistent part segmentations in a self-supervised manner.
Somewhat similar to our objective, class activation maps (CAMs) based methods~\cite{zhou2016learning,selvaraju2017grad} propose to localize the dense response on image with respect to a trained classifier.
However, without a learned part classifier, CAMs cannot be directly applied to our problem setting.
Recently, Collins~\etal~\cite{collins2018deep} propose deep feature factorization (DFF) to estimate the common part segments in images through Non-negative Matrix Factorization (NMF)~\cite{lee1999learning} on the ImageNet CNN features.
%
However, DFF requires joint optimization during inference time, and it is costly to impose other constraints or loss functions on part maps since there is no standalone inference module.
By posing as neural network inference, the proposed SCOPS can readily leverage the wealth of neural network loss functions developed in recent years. 
Any additional constraints can be jointly optimized during training time on large scale datasets, and the trained segmentation network could be applied on a single image during inference.


\vspace{1mm}
\noindent \textbf{Landmark detection}
Recently, several techniques have been proposed to learn landmarks with weak or no supervision. Most of these works rely on geometric constrains and landmarks equivariance to transformations. Thewlis \etal~\cite{thewlis2017unsupervised} relied on geometric priors to learn landmarks that are invariant to affine and spline transformations.
Zhang~\etal~\cite{zhang2018unsupervised} added reconstruction loss by reconstructing a given input image with predicted landmarks and local features.~Honari \etal~\cite{honari2018improving} used a subset of labeled images and sequential multitasking to improve final landmark estimation.  
Simon~\etal~\cite{simon2017hand} used multi-view bootstrapping to improve accuracy of hand landmark estimation.~Suwajanakorn~\etal~\cite{suwajanakorn2018discovery} used multiple geometry aware losses to discover 3D landmarks.
In order to obtain unsupervised landmarks, most of these works rely on simplified
problem settings such as using cropped images with only a single object instance per image and allowing only minor occlusions. 
We aim to predict part segments which provide richer 
representation of objects compared to landmarks.

\vspace{1mm}
\noindent \textbf{Dense image alignment}
Part segmentation is also related to the task of dense alignment, where the objective is to densely match pixels or landmarks from an object to another object instance.
While conventional approaches utilize off-the-shelf feature descriptors matching to tackle the problem, \eg SIFT flow based methods~\cite{liu2011sift,kim2013deformable,bristow2015dense}, recent works~\cite{han2017scnet,zagoruyko2015learning,zbontar2015computing,zbontar2016stereo,ham2016proposal} utilize annotated landmark pairs and deep neural networks to learn a better feature descriptor or matching function.
To avoid the cost of dense annotation, recent works propose to learn the dense alignment under the weakly supervised setting where only image pairs are required.
Rocco~\etal~\cite{rocco2017convolutional,rocco2018end} propose to jointly train the feature descriptor and spatial transformation by maximizing the inlier count, while Shu~\etal~\cite{shu2018deforming} propose Deforming Autoencoder to align faces and disentangle expressions.
However, these weakly supervised methods assume a certain family of spatial transforms, \eg affine or thin plate spline grid, to align objects with similar poses.
We argue that part segmentation is a more natural representation for semantic correspondences since matching each pixel between different instances would be an ill-posed problem. 
Part segmentations can also provide complex object deformations without heavily parameterized spatial transforms.

\vspace{1mm}
\noindent \textbf{Image co-segmentation}
Co-segmentation approaches predict the foreground pixels of the specific object given an image collection.
Most existing works~\cite{krause2015fine,rubinstein2013unsupervised,joulin2012multi,tsai2016semantic,rubio2012unsupervised} jointly consider all images within the collection to generate the final foreground segments via energy maximization, and thus not suitable for testing on standalone images.
In contrast, we propose an end-to-end trainable network that takes single image as input and outputs part segmentation which is more challenging but provides more information compared to foreground segmentation.

\begin{figure}[t!]
    \centering
    \includegraphics[width=1.0\linewidth]{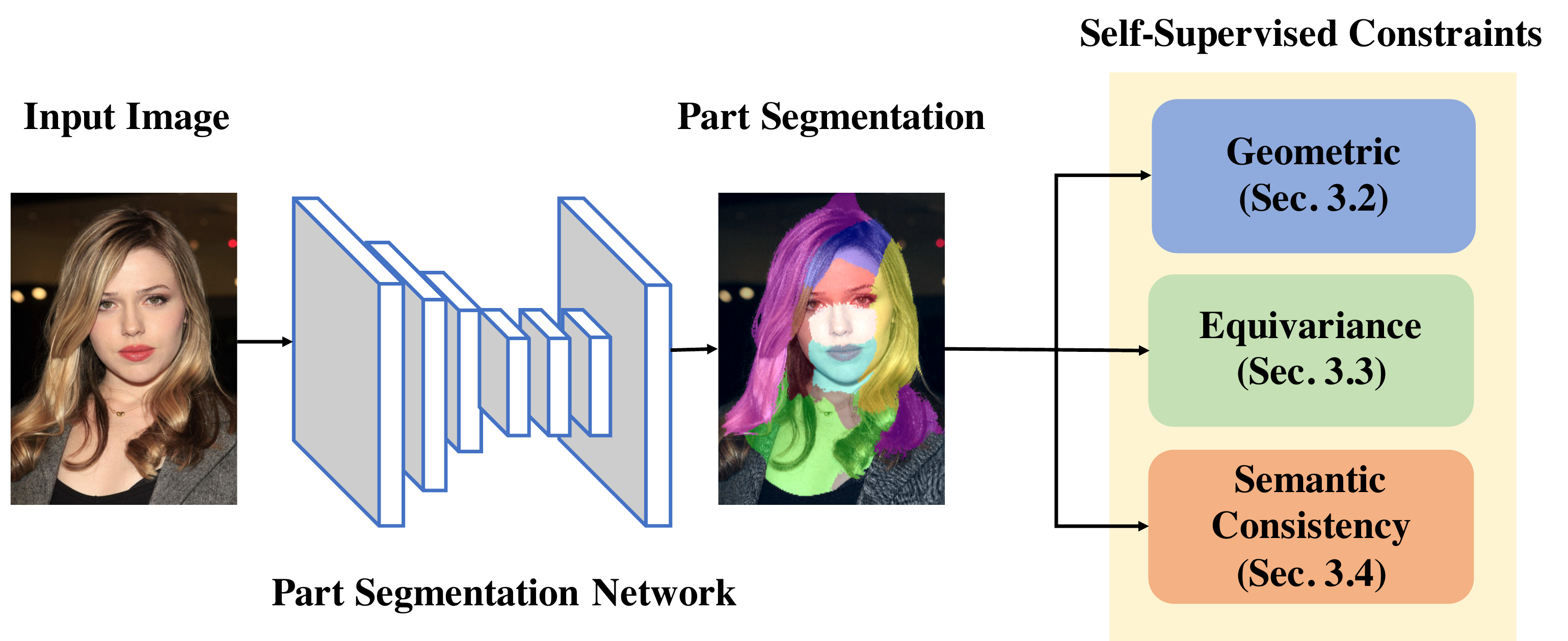}
    \mycaption{SCOPS framework}{Our network takes single image
    as input and predicts part segmentation. Geometric, Equivariance and Semantic Consistency constraints are used
    to train the network in a self-supervised manner.}
    \label{fig:framework}
    \vspace{-3mm}
\end{figure}
\vspace{-3mm}

\section{Self-Supervised Co-Part Segmentation}
\label{sec:methods}
\vspace{-0mm}

Given an image collection of the same object category, we aim to learn a deep neural network that takes a single 
image as input and outputs part segmentations. 
As outlined in Section~\ref{sec:intro},
we focus on the important characteristics of part segmentation
and devise loss functions that endorse these properties:
geometric concentration, robustness to variations, semantic
consistency, and objects as the union of parts. Here, we first describe
our overall framework followed by the description of different
loss functions and how they encourage the above-mentioned properties.
Along the way, we also comment on how our loss
functions are related and different to existing loss functions in the literature.

\vspace{-2mm}
\subsection{Overall Framework}
\vspace{-0mm}
Figure~\ref{fig:framework} shows the overall framework of our proposed method.
Given an image collection $\{\mathbf{I}\}$ of the same object category, 
we train a part segmentation network $\mathcal{F}$ parameterized by $\theta_f$, which is a fully convolutional neural network (FCN~\cite{fcn_pami}) with a channel-wise softmax layer in the end, 
to generate the part response maps $\mathbf{R} = \mathcal{F}(\mathbf{I} ; \theta_f) \in [0,1]^{(K+1) \times H \times W}$,
where $K$ denotes the number of parts and $H \times W$ is the image resolution. Our network predicts $K+1$ channels with an
additional channel indicating the background.
To obtain the final part segmentation results, we first normalize each part map with it maximum response value in the spatial dimensions $\mathbf{\hat{R}}(k,i,j) = \mathbf{R}(k,i,j) / \max_{u,v}(\mathbf{R}(k,u,v))$,
and we set the background map as constant with value $0.1$.
The purpose of this normalization is to enhance weak part responses.
Then the part segmentation is obtained with the $\arg\max$ function along the channel dimension.
We use DeepLab-V2~\cite{deeplab} with ResNet50~\cite{he2016deep} as our part segmentation network.

Since we do not assume the availability of any ground truth segmentation annotations,
we formulate several constraints as differentiable loss functions to encourage the above mentioned desired properties of a part segmentation, such as geometry concentration and semantic consistency.
The overall loss function for part segmentation network is a weighted sum of different loss functions which we describe next.
Contrary to several co-segmentation approaches~\cite{krause2015fine,rubinstein2013unsupervised,joulin2012multi,tsai2016semantic,rubio2012unsupervised}, 
which require multiple images during test-time inference, 
our network only takes a single image as input during the test time resulting in better portability of our trained model to unseen test images.

\subsection{Geometric Concentration Loss}
\vspace{-1mm}

Pixels belonging to the same object part are usually spatially concentrated within an image and form a connected component unless there are occlusions or multiple instances.
To this end, we first impose the geometric concentration on the part response maps to shape the part segments.
Specifically, we 
utilize a loss term that encourages
all the pixels belonging to a part to be spatially close to
the part center.
The part center for a part $k$ along axis $u$ is calculated as
\begin{equation}
    c_u^k = \sum_{u,v} u \cdot \mathbf{R}(k,u,v)/z_k,
    \label{eq:centroid}
\end{equation}
where $z_k = \sum_{u,v} \mathbf{R}(k,u,v)$ is the normalization term to transform the part response map into a spatial probability distribution function.
Then, we formulate the geometric concentration loss as
\begin{equation}
    \mathcal{L}_{con} = \sum_{k} \sum_{u,v} ||\langle u,v \rangle - \langle c_u^k, c_v^k \rangle||^2 \cdot \mathbf{R}(k,u,v)/z_k,
\end{equation}
and it is differentiable with respect to $c_u^k$, $\mathbf{R}(k,u,v)$, and $z_k$.
This loss function encourages
geometric concentration of parts and 
tries to minimize the variance of spatial probability distribution function $\mathbf{R}(k,u,v)/z_k$.
This loss is closely related to
ones used in recent unsupervised landmark estimation 
techniques~\cite{zhang2018unsupervised,thewlis2017unsupervised}. While Zhang~\etal~\cite{zhang2018unsupervised} approximate the landmark response maps with Gaussian distributions, 
we apply concentration loss mainly for penalizing part responses away from the part center.

Besides concentration loss, \cite{zhang2018unsupervised} and \cite{thewlis2017unsupervised} propose a form of separation (diversity) loss that maximizes the distance between different landmarks.
However, we do not employ such constraint as this constraint
would results in separated part segments with background
pixels in between.

\begin{figure}[t!]
    \centering
    \includegraphics[width=1.0\linewidth]{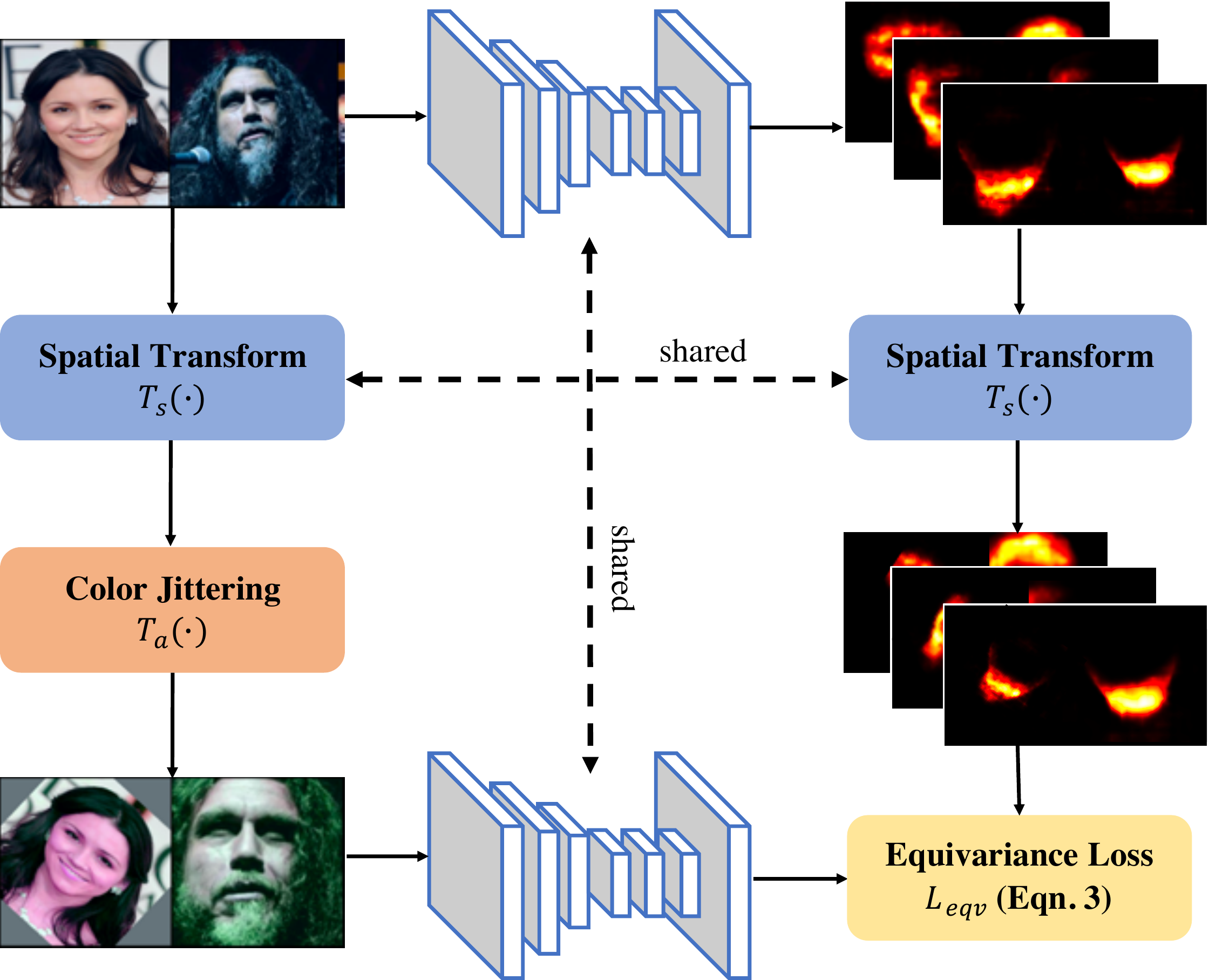}
    \mycaption{Equivariance loss}{We transform a given image
    with a random spatial transform and color jittering. We also
    transform the part segmentation of the given image using the
    same spatial transform to compare against the part segmentation
    of the transformed image via equivariance loss.}
    \label{fig:eqv}
    \vspace{-4mm}
\end{figure}

\subsection{Equivariance Loss}
\vspace{-1mm}

The second property that we want to advocate is that
part segmentation should be robust to the appearance and pose variations.
Figure~\ref{fig:eqv} illustrates how we employ the equivariance constraints to encourage the robustness to variations.
For each training image, we draw a random spatial transform $T_s(\cdot)$ and appearance perturbation $T_a(\cdot)$ from a predefined parameter range.
The detailed transform parameters are present in the supplementary material.
%
Then we pass both the input image $I$ and transformed image $I'=T_s(T_a(I))$ through the segmentation network and obtain the corresponding response maps $\mathbf{R}$ and $\mathbf{R}'$.
Given these part response maps, we compute the part centers
$\langle c_u^k, c_v^k \rangle$ and $\langle c_u^{k'}, c_v^{k'}\rangle$
using Eqn.~\ref{eq:centroid}.
Then, the equivariance loss is defined as
\begin{align}
    \mathcal{L}_{eqv} &= \lambda_{eqv}^s D_{KL}(\mathbf{R}'||T_s(\mathbf{R})) \nonumber \\
    &+ \lambda_{eqv}^c \sum_{k} ||\langle c_u^{k'}, c_v^{k'} \rangle - T_s(\langle c_u^k, c_v^k \rangle)||^2,
    \vspace{-2mm}
\end{align}
where $D_{KL}(\cdot)$ is the Kullback--Leibler divergence distance, and $\lambda_{eqv}^s, \lambda_{eqv}^c$ are the loss balancing coefficients.
The first term corresponds to the part segmentation equivariance, and the second term denotes the part center equivariance.
We use random similarity transformations (scale, rotation, and shifting) for spatial transforms.
We also experimented with more complex transformations such as 
projective and thin-plate-spline transformations, but did not
observe any improvements in part segmentation.

Recent works on unsupervised landmark estimation~\cite{zhang2018unsupervised,thewlis2017unsupervised} use
the above-mentioned equivariance loss on landmarks (part centers).
In this work, we extend the equivariance loss to part segmentation, and our experiments indicate that using only
equivariance on part centers is not sufficient to obtain good
part segmentation results. 

\subsection{Semantic Consistency Loss}
\vspace{-1mm}

\begin{figure}[t!]
    \centering
    \includegraphics[width=0.9\linewidth]{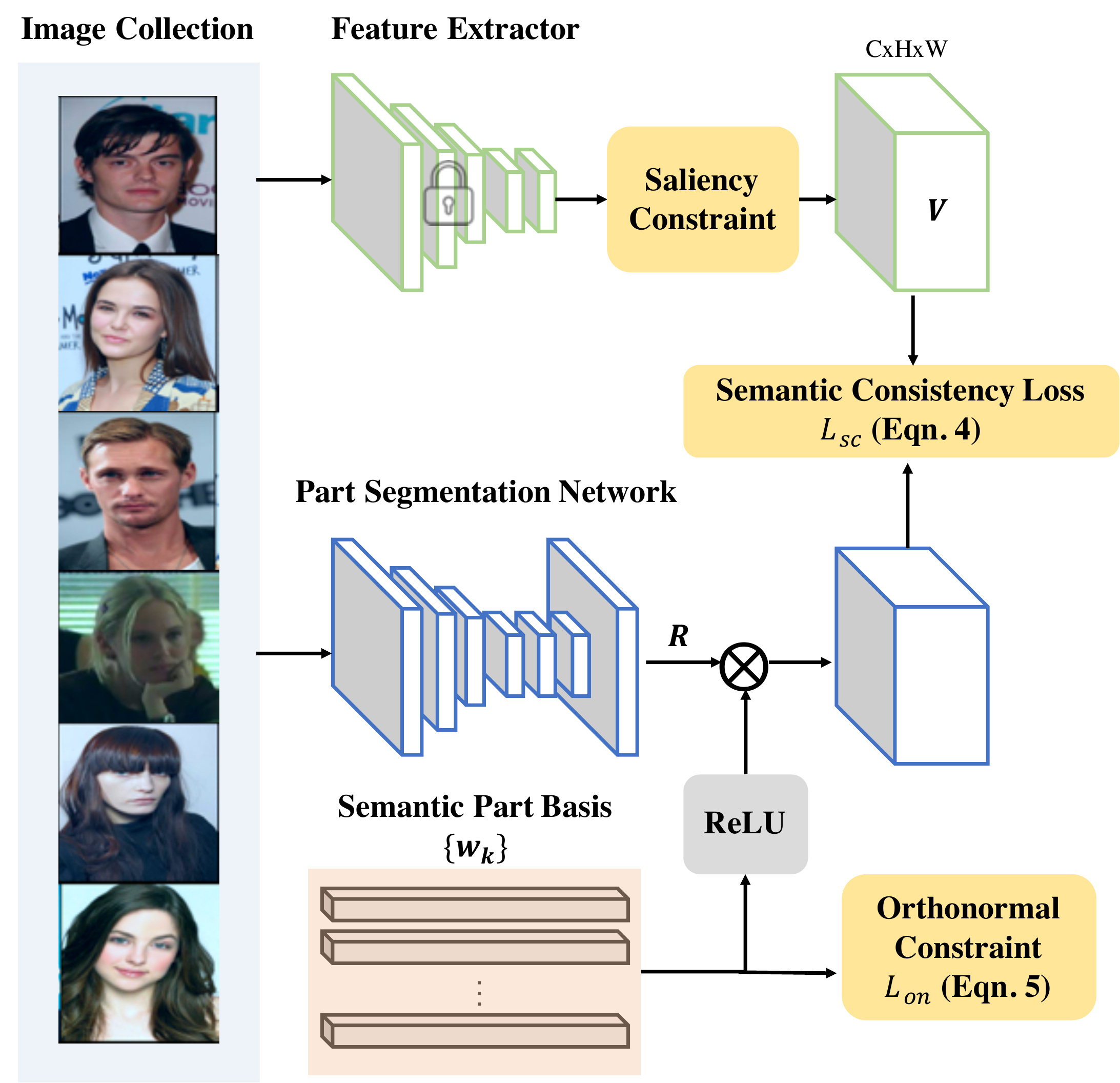}
    \mycaption{Semantic consistency loss}{We enforce semantic consistency of parts across instances by learning a semantic part basis that is shared across all images. 
    We use orthonormal constraint to learn distinct part basis, and we use saliency constraint to encourage parts to appear on foreground objects.}
    \label{fig:sc}
    \vspace{-3mm}
\end{figure}

Although equivariance loss favor part segmentations that are robust to some object variations,
the synthetically created transformations would not be sufficient 
to produce consistency across different instances since the appearance and pose variations between images are too high to be
modeled by any artificial transformations 
(See Figures~\ref{fig:teaser} and~\ref{fig:sc} for some sample instances).~To encourage semantic consistency across different object instances, 
we would need to explicitly leverage different instances in our loss function.

A key observation that we make use of is that the information about objects and parts is embedded in intermediate CNN features of classification networks ~\cite{bau2017network,gonzalez2018semantic,collins2018deep}. 
We devise a novel semantic consistency 
loss function that taps into this hidden part information
of ImageNet trained features~\cite{Krizhevsky_NIPS_2012,vgg,he2016deep}, which are readily available these days.
Following the observation in~\cite{collins2018deep}, 
we assume that we can find representative feature clusters in the given
classification features that are corresponding to different part segments. 

Formally, given $C$-dimensional classification
features $\mathbf{V} \in \mathcal{R}^{C\times H\times W}$,
we like to find $K$ representative part features 
$\mathbf{w}_k \in \mathcal{R}^C, k \in \{1,2,..., K\}$.
We simultaneously learn part segmentation $\mathbf{R}$
and these representative part features $\{\mathbf{w}_k\}$
such that the classification features $\mathbf{V}(u,v)$ 
of an $(u,v)$ pixel belonging to $k^{th}$ part is 
close to $\mathbf{w}_k$ \ie $||\mathbf{V}(u,v) - \mathbf{w}_k||^2 \rightarrow 0$.
Since the number of parts $K$ is usually smaller than
feature dimensionality $C$, we can see the representative
part features $\{\mathbf{w}_k\}$ as spanning a $K$-dimensional subspace in a $C$-dimensional space. We call these representative part features as part basis vectors.

Figure~\ref{fig:sc} illustrates the semantic consistency loss.
Given an image $\mathbf{I}$, we obtain its part response map $\mathbf{R}$.
We also pass $\mathbf{I}$ into a pre-trained classification network and obtain feature maps of an intermediate CNN layer.
The feature map is bi-linearly up-sampled to have the same spatial resolution of $\mathbf{I}$ and $\mathbf{R}$, resulting
in $\mathbf{V} \in \mathcal{R}^{C\times H\times W}$. 
We learn a set of part basis vectors $\{\mathbf{w}_k\}$ that
are globally shared across different object instances (training
images) using the following semantic consistency loss:
\begin{equation}
\vspace{-1mm}
    \mathcal{L}_{sc} = \sum_{u,v} || \mathbf{V}(u,v) -  \sum_{k} \mathbf{R}(k,u,v)\mathbf{w}_k ||^2, 
    \label{eq:sc}
\vspace{-1mm}
\end{equation}
where $\mathbf{V}(u,v) \in \mathcal{R}^C$ is the feature vector sampled at spatial location $(u,v)$.
We learn both the part segmentation $\mathbf{R}$ and the basis
vectors $\{\mathbf{w}_k\}$ at the same time using standard
back-propagation. To ensure that different part basis
vectors do not cancel each other out, we enforce non-negativity
on both features $\mathbf{V}$ and basis vectors $\{\mathbf{w}_k\}$ by passing them through a ReLU layer.
The part segmentation $\mathbf{R}$ is naturally non-negative as it is the output of a softmax function.

We view the semantic consistency loss as a linear subspace recovery problem with respect to the embedding space provided
by the feature extractor on the input image collection.
As the training progresses, the part bases can gradually converge to the most representative direction of each part in the embedding space provided by the pre-trained deep features, and the recovered subspace can be described as the span of the basis $\{\mathbf{w}_k\}$.
Furthermore, the non-negativity ensures that the weights $\mathbf{R}(k,u,v)$ could be interpreted as part responses.
With the proposed semantic consistency loss, we explicitly enforce the cross-instance semantic consistency through the learned part basis $\{\mathbf{w}_k\}$ since the same part response would have similar semantic feature embedding in the pre-trained feature space.

 \vspace{1mm}
 \noindent \textbf{Orthonormal Constraint}
When training with the semantic consistency loss, it is possible for the different basis to have similar feature embedding, especially when $K$ is large or the underlying rank of the subspace is smaller than $K$.
Having similar part basis, the part segmentation could become noisy since the response from multiple channels could all represent the same part segment. 
Therefore, we propose to impose an additional orthonormal
constraint on the part basis $\mathbf{w}_k$ to push the part bases apart.
Let $\hat{\mathbf{W}}$ denotes the matrix with each row as a normalized part basis vector $\hat{\mathbf{w}_k} = \mathbf{w}_k/||\mathbf{w}_k||$, and we formulate the orthonormal constraint as a loss function on $W$: 
\begin{equation}
\vspace{-1mm}
    \mathcal{L}_{on} = ||\hat{\mathbf{W}}\hat{\mathbf{W}}^T -\mathbb{I}_K||_F^2,
\vspace{-1mm}
\end{equation}
where $||\cdot||_F^2$ is Frobenius norm and $\mathbb{I}_K$ is the identity matrix of size $K\times K$.
The idea is to minimize the correlation between different basis vectors, and thus we can obtain a more concise basis set resulting in better part responses.

\vspace{1mm}
\noindent \textbf{Saliency Constraint}
We observe that, when the input image collection is small, or the number of parts $K$ is large, the proposed method tends to pick up some common background regions as object parts.
%
%
To tackle this issue, we utilize an unsupervised saliency detection method~\cite{zhu2014saliency} to suppress the background features in $\mathbf{V}$ so
that the learned part basis do not correspond to background regions.
To this end, for a given image and the unsupervised saliency map $\mathbf{D} \in [0,1]^{H \times W}$, we soft-mask the feature map $\mathbf{V}$ as $\mathbf{D}\circ\mathbf{V}$, where $\circ$ is the Hadamard (entry-wise) product, before passing it into the semantic consistency loss function.
Considering the non-salient pixels where $\mathbf{D} (u,v) = 0$,
the semantic consistency loss (Eqn.~\ref{eq:sc}) can be
interpreted as solving the following equation:
\vspace{-1mm}
\begin{equation}
   \mathbf{R}(k,u,v)\mathbf{w}_k = 0,
 \vspace{-1mm}
\end{equation}
which is essentially projecting the non-salient background 
regions into the \emph{null space} of the learned subspace spanned by $\{\mathbf{w}_k\}$.
This saliency constraint encapsulates our prior knowledge that
parts appear on objects (not background) and the union of parts
forms an object.
Several co-segmentation techniques~\cite{rubinstein2013unsupervised,chang2011co,fu2015object} also make use of
saliency maps to improve the segmentation result.
However, to our best knowledge, we are the first work to impose the saliency constraint in the feature reconstruction loss.

Related to our semantic consistency loss, a recent work~\cite{collins2018deep} proposed a deep feature factorization (DFF) technique for part discovery. Instead of learning a part basis,
DFF proposes to directly factorize features 
$\mathbf{V}$ into response maps $\mathbf{R}$ and
basis matrix $\mathbf{W}$ using non-negative matrix factorization 
(NMF); $\mathbf{V} \rightarrow \mathbf{R} \mathbf{W}$.
Although DFF alleviates the need for learning
a part basis and also training segmentation network, our learning strategy has several advantages compared to DFF.
First, we can make use of mini-batches and standard
gradient descent optimization techniques for learning part basis, whereas DFF performs NMF over the entire image collection at once during inference time. 
This makes our learning technique scalable to learning on large image collections and can be applied on single test image. 
Second, learning the part segmentation and basis using neural networks enables
easy incorporation of different constraints on the part basis 
(\eg orthonormal constraint) as well as the incorporation of
other loss functions such as concentration and equivariance. 
Our experiments indicate that these loss functions are essential to obtain good
part segmentations that are semantically consistent across images. 

\vspace{-2mm}
\section{Experiments}
\label{sec:experiments}
\vspace{-1mm}

Throughout the experiments, we refer to our technique as ``SCOPS" 
(Self-supervised Co-Part Segmentation).~Since SCOPS is self-supervised, the segmentation does not necessarily correspond to the human annotated object parts.
Therefore, we quantitatively evaluate SCOPS
with two different proxy measures on 
different object categories, 
including CelebA~\cite{liu2015faceattributes}, AFLW~\cite{koestinger2011annotated} (human faces),
CUB~\cite{wah2011caltech} (birds), and PASCAL~\cite{PASCAL_VOC_2010} (common objects) datasets.

On CelebA, AFLW, and CUB datasets,
we convert our part segmentation into landmarks by taking part centers (Eqn.~\ref{eq:centroid}) and evaluate against groundtruth annotations.
Following recent works~\cite{zhang2018unsupervised,thewlis2017unsupervised}, we fit a linear regressor that learns to map
the detected landmarks to groundtruth landmarks and evaluate
the resulting model on test data.
On PASCAL, we aggregate the part segmentations and evaluate them with the foreground segmentation IOU.

\vspace{1mm}
\paragrapht{Implementation Details}
We implement SCOPS\footnote{The code and models are available at \url{https://varunjampani.github.io/scops}} with PyTorch, and we train the networks with a single Nvidia GPU.
We use \texttt{relu5\_2} concatenated with \texttt{relu5\_4} from VGG-19~\cite{vgg} as the pre-trained features $\mathbf{V}$ for the semantic consistency loss. 
%



\begin{table} [t]
	\mycaption{Landmark evaluation on unaligned CelebA}{
	Mean L2 distance 
	comparing SCOPS to recent works (left) and also ablation with different
	loss functions (right).}
	\label{table:celebA}
	\footnotesize
	\begin{minipage}{.25\linewidth}
	\begin{tabular}{>{\raggedright\arraybackslash}p{2.3cm}>{\centering\arraybackslash}p{1.2cm}}
		\toprule
		\textbf{Method} & \textbf{Error} (\%) \\
		\midrule
		ULD (K=8)~\cite{zhang2018unsupervised,thewlis2017unsupervised} & 40.82  \\
		DFF (K=8)~\cite{collins2018deep} & 31.30 \\
		\cmidrule{1-2}
		SCOPS (K=4) & 21.76  \\
		SCOPS (K=8) & 15.01 \\
		\bottomrule
	\end{tabular}
	\end{minipage}%
	\hspace{2.3cm}
	\begin{minipage}{.1\linewidth}
	\begin{tabular}{lc}
		\toprule
		 \textbf{SCOPS}(K=8) & \textbf{Error} (\%) \\
		\midrule
		only $\mathcal{L}_{sc}$ & 23.53 \\
		w/o $\mathcal{L}_{sc}$ & 28.49 \\
		w/o $\mathcal{L}_{con}$ & 21.85 \\
		w/o $\mathcal{L}_{eqv}$ & 18.60 \\
		w/o Saliency & 22.11 \\
		\bottomrule
	\end{tabular}
	\end{minipage}
	\vspace{-1mm}
\end{table}

\begin{table} [t]
    \vspace{-1mm}
    \mycaption{Landmark evaluation on unaligned AFLW}{
    Mean L2 distance 
	comparing SCOPS to recent works.}
    \label{table:aflw}
    \centering
    \footnotesize
    
    \begin{tabu}{lccc}
        \toprule
        Method & ULD (K=8)~\cite{zhang2018unsupervised,thewlis2017unsupervised} & DFF (K=8)~\cite{collins2018deep} & SCOPS (K=8)\\
        \midrule
        Error (\%) & 25.03 & 20.42 & 16.54\\

        \bottomrule
    \end{tabu}
    \vspace{-1mm}
\end{table}

\vspace{-2mm}
\subsection{Faces from Unaligned CelebA/AFLW}
\label{subsec:celeba}
\vspace{-2mm}

The CelebA dataset contains around 20k face images, each annotated with a tight bounding box around face and 5 facial landmarks.
One of the main advantages of SCOPS is that it is relatively robust to pose and
viewpoint variations compared to recent landmark estimation works~\cite{zhang2018unsupervised,thewlis2017unsupervised}.
To demonstrate this, we experiment with \emph{unaligned}
CelebA images where we choose images with face covering more than 30\% of the pixel area.
Following the settings in~\cite{zhang2018unsupervised}, we also exclude 
MAFL~\cite{zhang2014facial} (subset of CelebA) test images from the train set resulting in a total of 45609 images.
We use the MAFL train set (5379 images) to fit the linear regression model and test on the MAFL test set (283 images).

In Table~\ref{table:celebA}, we report the landmark regression errors in terms of mean L2 distance normalized by inter-ocular distance.
To compare with the existing unsupervised landmark discovery works, we implement the loss functions, including concentration, separation, landmark equivariance, and reconstruction, as proposed in \cite{zhang2018unsupervised} and \cite{thewlis2017unsupervised}.
We train our base network with these constraints and refer it as ``ULD".
To validate our implementation of ULD, 
we train it on the align celebA images that yields 5.42\% landmark estimation 
error, which is comparable to the reported 5.83\% in \cite{thewlis2017unsupervised} and 3.46\% in \cite{zhang2018unsupervised}.
However, when training and testing with the unaligned images, we found that ULD has difficulty in converging to semantically meaningful landmark locations, resulting in high errors.
We also compare with a recent self-supervised part segmentation technique of DFF~\cite{collins2018deep} by considering the part responses as landmark detections.
We train SCOPS to predict 4 and 8 parts with all the proposed constraints and show the comparison results in Table~\ref{table:celebA} (left).
The results show that SCOPS performs favorably against other methods.
The visual results of SCOPS ($K=8$) in Fig.~\ref{fig:ablation} shows that SCOPS part segments are more semantically consistent across different images compared to existing techniques.
In addition, we train SCOPS on the AFLW dataset~\cite{koestinger2011annotated}, which contains 4198 face images (after filtering) with 21 annotated landmarks. Following \cite{zhang2018unsupervised}, we pretrain the model on CelebA and finetune on AFLW.
We show the results in Table~\ref{table:aflw}. Results indicate that SCOPS outperforms both ULD and DFF on this dataset images as well.
Even though the landmark prediction accuracy do not directly measure the learned part segmentation quality, these results demonstrate that the learned part segmentations are semantically consistent across instances under the challenging unaligned setting.

\vspace{1mm}
\newlength{\ablationw}
\setlength{\ablationw}{0.22\linewidth}
\newlength{\ablationh}
\setlength{\ablationh}{0.22\linewidth}

\newcommand{\ablationrow}[2]{            
    \includegraphics[width=\ablationw,height=\ablationh]{figures/celebawild/#1/iter_#2/test/part_dcrf_overlay/009567.jpg} &
    \includegraphics[width=\ablationw,height=\ablationh]{figures/celebawild/#1/iter_#2/test/part_dcrf_overlay/071289.jpg} &
    \includegraphics[width=\ablationw,height=\ablationh]{figures/celebawild/#1/iter_#2/test/part_dcrf_overlay/108309.jpg} &
    \includegraphics[width=\ablationw,height=\ablationh]{figures/celebawild/#1/iter_#2/test/part_dcrf_overlay/164803.jpg}
}

\begin{figure}[ht]

    \centering
    \footnotesize
    \begin{tabular}
    {   b{4mm}
        @{\hspace{0mm}}c@{\hspace{0.3mm}} 
        @{\hspace{0mm}}c@{\hspace{0.3mm}}
        @{\hspace{0mm}}c@{\hspace{0.3mm}} 
        @{\hspace{0mm}}c@{\hspace{0.3mm}} 
    }
        \vfill \rotatebox{90}{image} \vfill &
        \includegraphics[width=\ablationw,height=\ablationh]{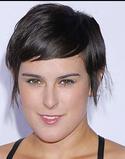} &
        \includegraphics[width=\ablationw,height=\ablationh]{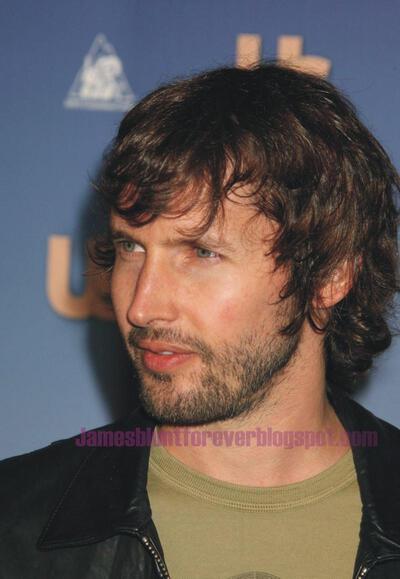} &
        \includegraphics[width=\ablationw,height=\ablationh]{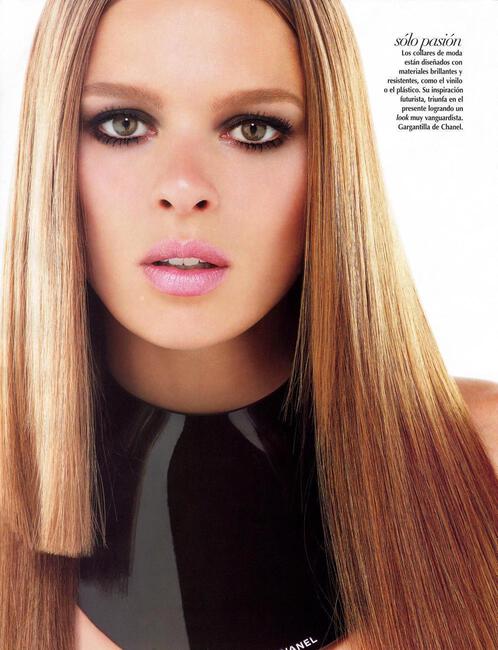} &
        \includegraphics[width=\ablationw,height=\ablationh]{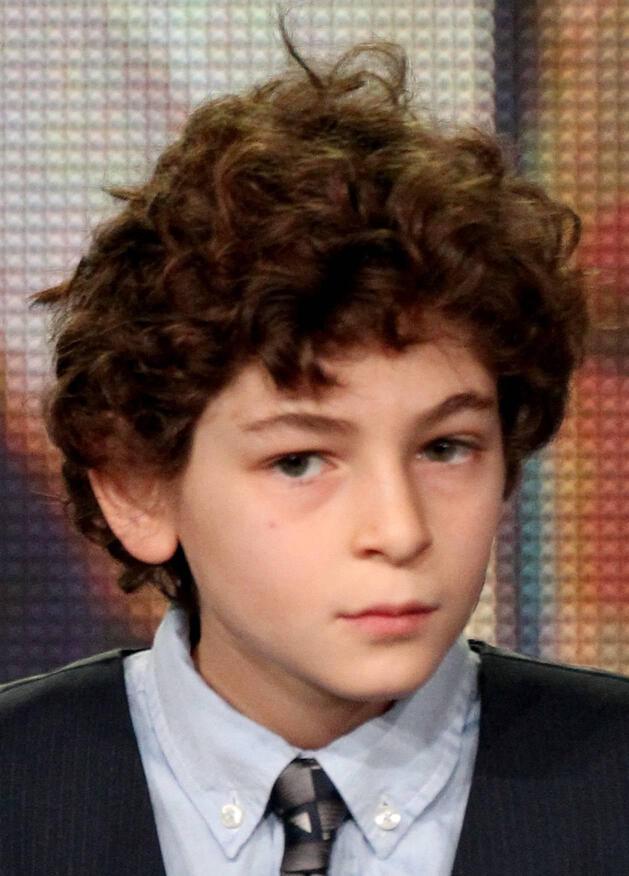} \\
        \midrule
            
        \vfill \rotatebox{90}{ ULD~\cite{zhang2018unsupervised,thewlis2017unsupervised} } \vfill &
        \ablationrow{1111_celebawild_filter30_k8_con1e1_sep1e2_lmeqv1e2}{100000}\\
    
        \vfill \rotatebox{90}{ DFF~\cite{collins2018deep} } \vfill &
        \includegraphics[width=\ablationw,height=\ablationh]{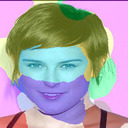} &
        \includegraphics[width=\ablationw,height=\ablationh]{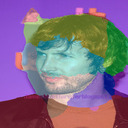} &
        \includegraphics[width=\ablationw,height=\ablationh]{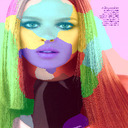} &
        \includegraphics[width=\ablationw,height=\ablationh]{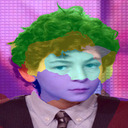}\\
        
        \midrule

        
        \vfill \rotatebox{90}{ SCOPS} \vfill &
        \ablationrow{1111_celebawild_filter30_dff_52_54_k8_dff1e4_con1e1_eqv1e3_lmeqv1e2}{90000}\\
        
        \midrule
        \vfill \rotatebox{90}{ w/o $\mathcal{L}_{sc}$ } \vfill &
        \ablationrow{1111_celebawild_filter30_dff_54_k8_dff0_con1e1_eqv1e3_lmeqv1e2}{100000}\\
        
        \vfill \rotatebox{90}{ only $\mathcal{L}_{sc}$} \vfill &
        \ablationrow{1111_celebawild_filter30_dff_54_k8_dff1e4_nosal}{100000}\\

        \vfill \rotatebox{90}{ w/o $\mathcal{L}_{con}$ } \vfill &
       \ablationrow{1111_celebawild_filter30_dff_54_k8_dff1e4_eqv1e3_lmeqv1e2}{100000}\\
        
        \vfill \rotatebox{90}{w/o $\mathcal{L}_{eqv}$ } \vfill &
        \ablationrow{1111_celebawild_filter30_dff_54_k8_dff1e4_con1e1}{100000}\\
        
        \vfill \rotatebox{90}{w/o saliency} \vfill &
        \ablationrow{1111_celebawild_filter30_dff_54_k8_dff1e4_con1e1_eqv1e3_lmeqv1e2_nosal}{100000}\\
        
    \end{tabular}
\mycaption{Visual results on CelebA face images}{SCOPS produce consistent part segments compared to existing techniques. Also shown
is the effect of different loss constraints.}
\label{fig:ablation}
\vspace{-3mm}
\end{figure}

\noindent \textbf{Ablation Study}
To validate the individual contribution of the different constraints, we conduct a detailed ablation study and show the results in Table~\ref{table:celebA} (right).
The corresponding visual results are shown in Figure~\ref{fig:ablation}.
While removing any of the constraints results in worse performance, the semantic consistency loss $\mathcal{L}_{sc}$ is the most important constraint in the proposed framework, and removing it would cause the most performance drop.
Visual results in Figure~\ref{fig:ablation} indicate that the learned parts would not have a semantic meaning
without $\mathcal{L}_{sc}$. Results also indicate that training without geometric concentration loss $\mathcal{L}_{con}$
would cause some parts dominating large image areas, and 
no equivariance loss $\mathcal{L}_{eqv}$ makes the learned parts not consistent across images.
These results demonstrate that all our loss functions are essential to learning good part segmentations.

\begin{table} [t!]
    \mycaption{Landmark evaluation on CUB}{
	Normalized L2 distance 
	comparing SCOPS to recent techniques (K=4).}
	\label{table:cub}
	\centering
	\footnotesize
	
	\begin{tabu}{lccc}
		\toprule
		Method & CUB-001 & CUB-002 & CUB-003\\
		\midrule
		ULD~\cite{zhang2018unsupervised,thewlis2017unsupervised} & 30.12 & 29.36 & 28.19 \\
		DFF~\cite{collins2018deep} & 22.42 & 21.62& 21.98\\
		\midrule
		SCOPS & 18.50& 18.82 & 21.07\\
		\bottomrule
	\end{tabu}
\end{table}

\newcommand{\cubrow}[2]{            
	\includegraphics[width=\ablationw,height=\ablationh]{figures/cub/002/#1/laysan_albatross_0102_611.#2} &
	\includegraphics[width=\ablationw,height=\ablationh]{figures/cub/002/#1/laysan_albatross_0082_524.#2} &
	\includegraphics[width=\ablationw,height=\ablationh]{figures/cub/002/#1/laysan_albatross_0039_924.#2} &
	\includegraphics[width=\ablationw,height=\ablationh]{figures/cub/002/#1/laysan_albatross_0006_702.#2}
}

\begin{figure}[t!]
	\centering
	\footnotesize
    \begin{tabular}
    {   b{4mm}
        @{\hspace{0mm}}c@{\hspace{0.3mm}} 
        @{\hspace{0mm}}c@{\hspace{0.3mm}}
        @{\hspace{0mm}}c@{\hspace{0.3mm}} 
        @{\hspace{0mm}}c@{\hspace{0.3mm}} 
    }

        \vfill \rotatebox{90}{ Image } \vfill &
        \cubrow{images}{jpg} \\
        \vfill \rotatebox{90}{ ULD~\cite{zhang2018unsupervised,thewlis2017unsupervised} } \vfill &
        \cubrow{uld}{jpg} \\
        \vfill \rotatebox{90}{ DFF~\cite{collins2018deep} } \vfill &
        \cubrow{dff}{jpg} \\
        \vfill \rotatebox{90}{ SCOPS } \vfill &
        \cubrow{scops}{jpg} \\

    \end{tabular}

\mycaption{Visual results on CUB bird images}{SCOPS is robust to pose and camera variations while having better boundary adherence compared to other techniques.}
\label{fig:cub}
\vspace{-4mm}
\end{figure}

\vspace{-1mm}
\subsection{Birds from CUB}
\vspace{-1mm}

We also evaluate the proposed method on a more challenging bird images from CUB-2011 dataset~\cite{wah2011caltech}, which consists of 11,788 images with 200 categories of birds and 15 landmark annotations.
The dataset is challenging because of the various bird poses, e.g., standing, swimming, or flying, as well as the different camera viewpoints.
We train SCOPS with $K=4$ on first three bird categories and compare to ULD and DFF.
We show some qualitative results in Figure~\ref{fig:cub}.
With such level of object deformation, we found that ULD has difficulty in localizing meaningful parts.
Compared to DFF, the part segments produced by SCOPS has better boundary alignment within and outside the object, and the learned part segmentation is also more consistent across instances.
Similar to previous Section~\ref{subsec:celeba}, we use the landmark detection as the proxy task through considering the part centers as detected landmarks.
To account for the varying bird sizes across images, we normalize the landmark estimation error by the width and height of the provided ground truth bounding boxes. 
%
Table~\ref{table:cub} shows the quantitative results of different techniques.
For all the three bird categories, SCOPS performs favorably against both the ULD~\cite{zhang2018unsupervised,thewlis2017unsupervised}, and DFF~\cite{collins2018deep} techniques.
For the CUB-2011 dataset, SCOPS as well as other techniques
do not distinguish between left-right symmetric parts.
For instance, the left-wing and right-wing are often predicted as the same part.
From a part segmentation perspective, such behavior is reasonable.
However, considering the landmark regression task, the part center of the two fanned out wings would be on the main body, resulting in less meaningful landmarks.
As a result, the landmark regression error may not accurately reflect the co-part segmentation quality
and distinguishing the symmetric semantic parts remains a challenging problem on this dataset images.

\vspace{-1mm}
\subsection{Common Objects from PASCAL}
\vspace{-1mm}

We also apply SCOPS on the PASCAL VOC dataset~\cite{PASCAL_VOC_2010}, which contains images with common objects with various deformations, viewing angles, and occlusions.
We extract the images that contain the specific object
category while the object bounding box occupies at least 20\% of the whole image. 
To remove significant occlusions in the images, we further exclude the images where only a small portion of ground-truth parts are present in the PASCAL-part dataset~\cite{chen_cvpr14}.
The models are trained separately for each object category with $K=4$.
Although the PASCAL-part dataset~\cite{chen_cvpr14} provides the object parts annotation, a good self-supervised part segmentation can produce semantically consistent part segments that may not correspond to manually annotated part segments.
Therefore, we do not evaluate the results with the part-level Intersection over Union (IoU) since it is not a good indicator.
Instead, we evaluate the results as co-segmentation by aggregating the part segments and computing the foreground object segmentation IoU.
Since the co-segmentation metric only indicates the overall object localization and not the part segmentation consistency, 
this metric is only indicative of part segmentation quality.
We show some visual results in Figure~\ref{fig:horse} and the quantitative evaluations in Table~\ref{table:voc}.
In terms of IoU, SCOPS outperform DFF by a considerable margin, both with and without the CRF post-processing~\cite{krahenbuhl2011efficient}.
The visual results show that SCOPS is robust to various appearance and pose articulations.
%
We show additional visual results in the supplementary material.

\newlength{\horsew}
\setlength{\horsew}{0.22\linewidth}
\newlength{\horseh}
\setlength{\horseh}{0.22\linewidth}

\newcommand{\horserow}[2]{            
    \includegraphics[width=\horsew,height=\horseh]{figures/pascal/sheep/#1/2009_004945.#2} &
    \includegraphics[width=\horsew,height=\horseh]{figures/pascal/sheep/#1/2010_000189.#2} &
    \includegraphics[width=\horsew,height=\horseh]{figures/pascal/motor/#1/2008_004822.#2} &
    \includegraphics[width=\horsew,height=\horseh]{figures/pascal/motor/#1/2008_007054.#2}
}

\begin{figure}[t!]
	\centering
	\footnotesize
    \begin{tabular}
    {   b{4mm}
        @{\hspace{0mm}}c@{\hspace{0.3mm}} 
        @{\hspace{0mm}}c@{\hspace{0.3mm}}
        @{\hspace{0mm}}c@{\hspace{0.3mm}} 
        @{\hspace{0mm}}c@{\hspace{0.3mm}} 
        @{\hspace{0mm}}c@{\hspace{0.3mm}} 
    }

        \vfill \rotatebox{90}{ Image } \vfill &
        \horserow{img}{jpg} \\
        \vfill \rotatebox{90}{ SCOPS } \vfill &
        \horserow{scops}{jpg} \\

    \end{tabular}
\mycaption{Visual results on the PASCAL VOC datset~\cite{PASCAL_VOC_2010}}{SCOPS is robust to pose and appearance variations.}
\label{fig:horse}
\vspace{-0mm}
\end{figure}

\begin{table} [t]
    \vspace{-2mm}
    \scriptsize
    \mycaption{Evaluation on the PASCAL VOC dataset}{
    Cosegmentation IoU
    comparing SCOPS to DFF on 7 object classes of VOC (K=4).}
    \label{table:voc}
    \centering

    \begin{tabu}{lccccccc}
        \toprule
        class & horse & cow & sheep & aero & bus & car & motor\\
        \midrule
        DFF~\cite{collins2018deep}& 49.51 & 56.39 & 51.03 & 48.38 & 58.63 & 56.48 & 54.80 \\
        DFF+CRF~\cite{collins2018deep} &  50.96 & 57.64 & 52.29 & 50.87 & 58.64 & 57.56 & 55.86 \\
        \midrule
        SCOPS  &  55.76 & 60.79 & 56.95 & 69.02 & 73.82 & 65.18 & 58.53\\
        SCOPS+CRF &  \textbf{57.92} & \textbf{62.70} & \textbf{58.17} & \textbf{80.54} & \textbf{75.32} & \textbf{66.14} & \textbf{59.15} \\

        \bottomrule
    \end{tabu}
    \vspace{-4mm}
\end{table}

\vspace{-2mm}
\section{Concluding Remarks}
\label{sec:conclusion}
\vspace{-1mm}

We propose SCOPS, a self-supervised technique for co-part segmentation.
Given an image collection of an object category, SCOPS can learn to predict semantically consistent part segmentations without using any ground-truth annotations.
We devise several constraints, including geometric concentration, equivariance, as well as semantic consistency, to train a deep neural network to discover semantically consistent part segments while ensuring decent geometric configurations and cross instance correspondence.
Results on different types of image collections show that SCOPS is robust to different object appearances, camera viewpoints, as well as pose articulations. The qualitative and quantitative results show that SCOPS performs favorably against existing methods.
We hope that the proposed method could serve as a general framework for learning co-part segmentation.
%

\vspace{-3mm}
{\flushleft {\bf Acknowledgments.}}
W.-C. Hung is supported in part by the NSF CAREER Grant \#1149783, gifts from Adobe, Verisk, and NEC.

{\small
\bibliographystyle{ieee_fullname}
\bibliography{egbib}
}

\appendix
\onecolumn

\setlength{\ablationw}{0.11\linewidth}
\setlength{\ablationh}{0.11\linewidth}

\renewcommand{\ablationrow}[4]{            
    \includegraphics[width=\ablationw,height=\ablationh]{figures_supp_compressed/#1/iter_#2/web_html/images/#3/009567#4} &
    \includegraphics[width=\ablationw,height=\ablationh]{figures_supp_compressed/#1/iter_#2/web_html/images/#3/031034#4} &
    \includegraphics[width=\ablationw,height=\ablationh]{figures_supp_compressed/#1/iter_#2/web_html/images/#3/075297#4} &
    \includegraphics[width=\ablationw,height=\ablationh]{figures_supp_compressed/#1/iter_#2/web_html/images/#3/131414#4} &
    \includegraphics[width=\ablationw,height=\ablationh]{figures_supp_compressed/#1/iter_#2/web_html/images/#3/106570#4} &
    \includegraphics[width=\ablationw,height=\ablationh]{figures_supp_compressed/#1/iter_#2/web_html/images/#3/120303#4} &
    \includegraphics[width=\ablationw,height=\ablationh]{figures_supp_compressed/#1/iter_#2/web_html/images/#3/195686#4} &
    \includegraphics[width=\ablationw,height=\ablationh]{figures_supp_compressed/#1/iter_#2/web_html/images/#3/063020#4} 
}

\section{Optimization Objectives}

In the proposed method, we train the part segmentation network and the semantic part basis with several loss functions, including geometric concentration loss $\mathcal{L}_c$, equivariance loss $\mathcal{L}_{eqv}$ and semantic consistency loss $\mathcal{L}_{sc}$, with orthonomal constraint $\mathcal{L}_{ot}$.
The final objective function is a linear combination of these loss functions,
\begin{equation}
	\mathcal{L}_{all} = \lambda_c \mathcal{L}_c +  \lambda_{eqv} \mathcal{L}_{eqv} +  \lambda_{sc} \mathcal{L}_{sc} +  \lambda_{ot} \mathcal{L}_{ot} .
\end{equation}
In all our experiments, the weighting coefficients $(\lambda_c, \lambda_{eqv},\lambda_{sc},\lambda_{ot})$ are set to $(0.1,10,100,0.1)$.
We obtain these weights by conducting a coarse grid search on a subset of the CelebA dataset images~\cite{liu2015faceattributes}.

\section{Implementation Details of Equivariance Loss}
For spatial transform, we apply random $\pm 60 \degree$ rotation, $\pm 20\%$ shifting, $0.3x-2x$ scaling, 
as well as TPS transform with 5x5 grid and $\pm 10\%$ shifting for each control point.
For color transform, we apply jittering to brightness ($\pm30\%$), contrast ($\pm30\%$), saturation ($\pm20\%$), and hue ($\pm20\%$).
In our experiments, we find that the performance gain is robust with respect to a wide range of perturbation parameters.

\section{Additional Experimental Analysis}

\subsection{On Using Different Part Number $K$}
We evaluate the effects of selected part number $K$ on the unaligned CelebA dataset~\cite{liu2015faceattributes}.
We select $K=2,4,6,8,10$ and present the landmark estimation errors in Table~\ref{table:k}.
Note that with higher $K$ the landmarks' center are more robust to face pose variations and thus lead to lower error rate.
However, the performance seems to saturate after $K=8$.
The qualitative results in Figure~\ref{fig:k} also indicate the
similar trend. 
When observing the part segmentation results for $K=8$ and $K=10$, the additional parts seem to be not corresponding to any semantically meaningful area of the face.

 \begin{table} [h]
 	\mycaption{Landmark evaluation on CelebA with different K}{
 		Mean L2 distance.}
 	\label{table:k}
 	\centering
 	
 	\begin{tabu}{lccccc}
 		\toprule
 		SCOPS & K = 2 & K = 4 & K = 6 & K = 8 & K = 10\\
 		\midrule
 		Error (\%) & 25.46 & 21.76 & 15.92 & 15.01 & 14.71 \\
 		\bottomrule
 	\end{tabu}
 \end{table}

\subsection{Feature Visualization of Learned Semantic Part Basis}

\begin{figure*}[ht!]
	\centering
	
	\includegraphics[width=0.9\textwidth,height=0.9\textheight]{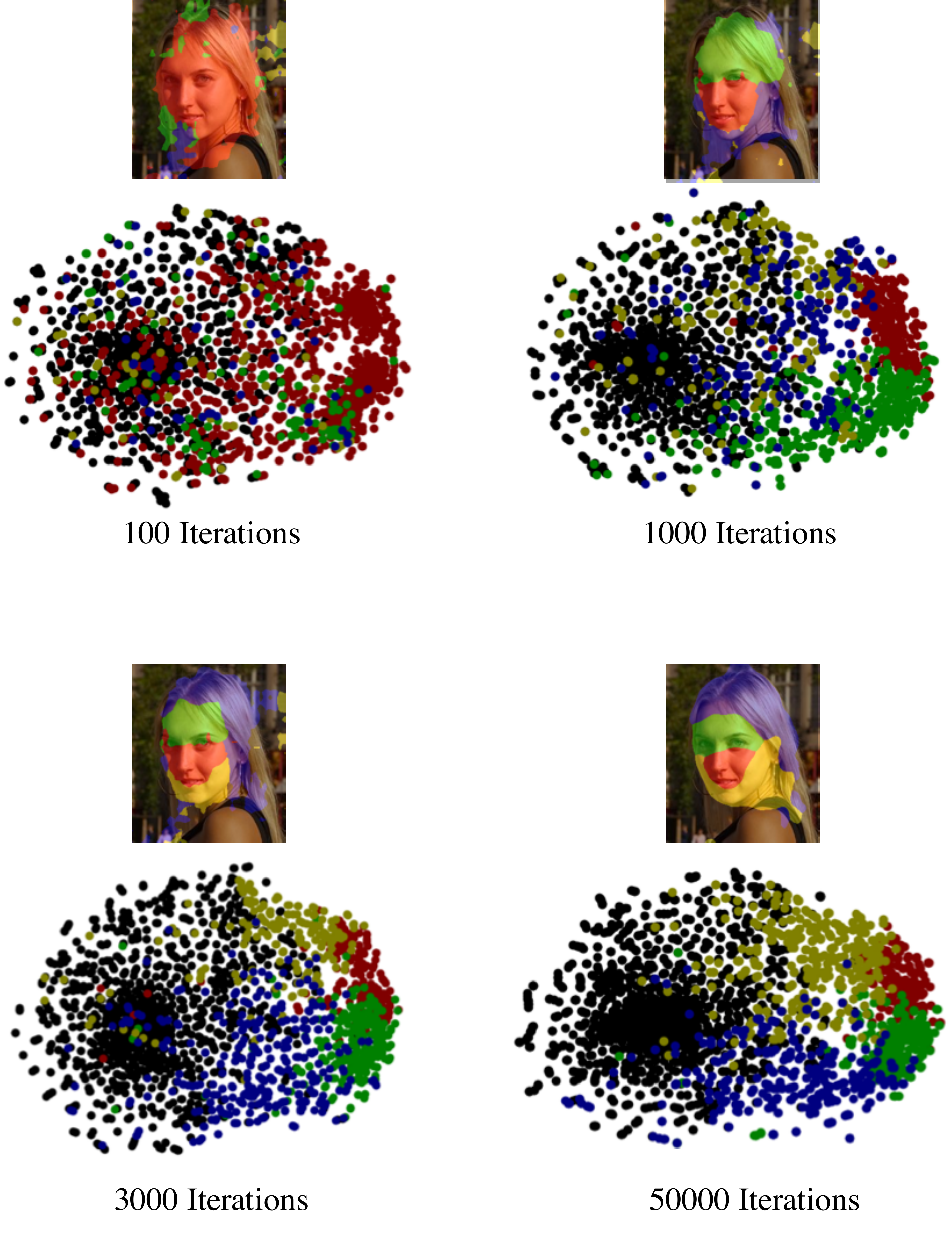}
	\mycaption{Training progression}{The top images show the part segmentation results while the bottom show the TSNE visualization on the imagenet feature of pixels and their corresponding segmentation class. Black dots are the background pixels.
    During the training process, we can see that the part segmentations are improving, while pixels with similar imagenet features are segmented as same part. 
    }
	\label{fig:tsne}
	
\end{figure*}

We also show how the learned part basis improves on training progress in Figure~\ref{fig:tsne}.
The top images show the part segmentation results while the bottom show the t-SNE visualization~\cite{maaten2008visualizing} on the ImageNet feature of pixels and their corresponding segmentation class i.e., the dot colors correspond to the part segmentation visualization colors, while the black dots represent the background pixels.
During the training process, we can see that the part segmentations are improving, while pixels with similar ImageNet features are segmented as same part.

\section{Quantitative Results on iCoseg}

We present additional evaluation on the iCoseg dataset~\cite{batra2010icoseg}, which is a commonly used co-segmentation dataset. Follow DFF~\cite{collins2018deep}, we select 5 image sets and train SCOPS with $K=4$ on the images. We aggregate the parts and evaluate with foreground segmentation IOU. The results show that SCOPS performs favorably against existing methods.

\begin{table} [h!]
    \vspace{-2mm}
    \mycaption{Evaluation on iCoseg}{
    Cosegmentation IoU
    comparing SCOPS to recent techniques on 5 image sets of iCoseg.}
    \label{table:icoseg}
    \centering
    
    \begin{tabu}{lccccc}
        \toprule
        Subset & Elephants & Taj-Mahal & Pyramids & Gymnastics1 & Statue of Liberty\\
        \midrule
        Rubinstein~\cite{rubinstein2013unsupervised} & 63 & 48 & 57 & \textbf{94} & \textbf{70} \\
        DFF~\cite{collins2018deep}& 65& 41 & 57 & 43 & 49 \\
        DFF-crf~\cite{collins2018deep}& 76 & 51 & 70 & 52 & 62 \\
        \midrule
        SCOPS  & 78.5 & \textbf{67.0} & 72.7 & 73.5 & 63.8 \\

        SCOPS-crf &  \textbf{81.5} & 66.0 & \textbf{75.0} & 81.5 & 65.7 \\
        \bottomrule
    \end{tabu}
    \vspace{-2mm}
\end{table}

\section{Qualitative Results on PASCAL Objects}

We present additional part segmentation visual results on different object class images obtained from the PASCAL dataset~\cite{PASCAL_VOC_2010}.
In Figures~\ref{fig:car}-\ref{fig:cow}, the results show that the proposed method works for both rigid (Figures~\ref{fig:car}-\ref{fig:motor}) and non-rigid (Figures~\ref{fig:sheep}-\ref{fig:cow}) objects, where SCOPS produce part segments that are consistent across different instances with variations in appearance, pose and camera viewpoints.

\section{Additional Results on Unaligned CelebA and CUB}

We provide more visual results on the unaligned CelebA dataset~\cite{liu2015faceattributes} and the CUB dataset~\cite{wah2011caltech} in Figures~\ref{fig:celeba}-\ref{fig:cub2}.
The results show that SCOPS is robust to pose and camera variations while having better boundary adherence compared to other techniques.

\renewcommand{\ablationrow}[4]{            
	\includegraphics[width=\ablationw,height=\ablationh]{figures_supp_compressed/#1/iter_#2/web_html/images/#3/189937#4} &
	\includegraphics[width=\ablationw,height=\ablationh]{figures_supp_compressed/#1/iter_#2/web_html/images/#3/197951#4} &
	\includegraphics[width=\ablationw,height=\ablationh]{figures_supp_compressed/#1/iter_#2/web_html/images/#3/134935#4} &
	\includegraphics[width=\ablationw,height=\ablationh]{figures_supp_compressed/#1/iter_#2/web_html/images/#3/177737#4} &
	\includegraphics[width=\ablationw,height=\ablationh]{figures_supp_compressed/#1/iter_#2/web_html/images/#3/102050#4} &
	\includegraphics[width=\ablationw,height=\ablationh]{figures_supp_compressed/#1/iter_#2/web_html/images/#3/080512#4} &
	\includegraphics[width=\ablationw,height=\ablationh]{figures_supp_compressed/#1/iter_#2/web_html/images/#3/147918#4} &
	\includegraphics[width=\ablationw,height=\ablationh]{figures_supp_compressed/#1/iter_#2/web_html/images/#3/176672#4} 
}

\begin{figure*}[ht!]
	\centering
	\footnotesize
	\begin{tabular}
		{   b{4mm}
			@{\hspace{0mm}}c@{\hspace{0.3mm}} 
			@{\hspace{0mm}}c@{\hspace{0.3mm}}
			@{\hspace{0mm}}c@{\hspace{0.3mm}} 
			@{\hspace{0mm}}c@{\hspace{0.3mm}} 
			@{\hspace{0mm}}c@{\hspace{0.3mm}} 
			@{\hspace{0mm}}c@{\hspace{0.3mm}} 
			@{\hspace{0mm}}c@{\hspace{0.3mm}} 
			@{\hspace{0mm}}c@{\hspace{0.3mm}} 
		}
		\vspace{3cm}
		\vfill \rotatebox{90}{Image} \vfill &
		\ablationrow{1111_celebawild_filter30_k8_con1e1_sep1e2_lmeqv1e2}{100000}{img}{.jpg}\\
		
		\vfill \rotatebox{90}{Saliency} \vfill &
		\ablationrow{saliency_rbd}{0}{saliency}{_rbd.jpg}\\
		
		\midrule
		
		\vfill \rotatebox{90}{ SCOPS (K=2)} \vfill &
		\ablationrow{1118_celebawild_filter30_dff_52_54_k2_dff1e4_con1e1_eqv1e3_lmeqv1e2}{50000}{partooverlaydcrf}{.jpg}\\
		
		\vfill \rotatebox{90}{ SCOPS (K=4)} \vfill &
		\ablationrow{1118_celebawild_filter30_dff_52_54_k4_dff1e4_con1e1_eqv1e3_lmeqv1e2}{50000}{partooverlaydcrf}{.jpg}\\
		
		\vfill \rotatebox{90}{ SCOPS (K=6)} \vfill &
		\ablationrow{1118_celebawild_filter30_dff_52_54_k6_dff1e4_con1e1_eqv1e3_lmeqv1e2}{50000}{partooverlaydcrf}{.jpg}\\
		
		\vfill \rotatebox{90}{ SCOPS (K=8)} \vfill &
		\ablationrow{1111_celebawild_filter30_dff_52_54_k8_dff1e4_con1e1_eqv1e3_lmeqv1e2}{90000}{partooverlaydcrf}{.jpg}\\
		
		\vfill \rotatebox{90}{ SCOPS (K=10)} \vfill &
		\ablationrow{1118_celebawild_filter30_dff_52_54_k10_dff1e4_con1e1_eqv1e3_lmeqv1e2}{50000}{partooverlaydcrf}{.jpg}\\

	\end{tabular}

	\mycaption{Effects of changing part number}{We apply SCOPS on the unaligned CelebA dataset with the part number K ranging from 2 to 10.}
	\label{fig:k}
	
\end{figure*}

\renewcommand{\ablationrow}[3]{            
    \includegraphics[width=\ablationw,height=\ablationh]{figures_supp_compressed/#1/#2/2008_000346#3} &
    \includegraphics[width=\ablationw,height=\ablationh]{figures_supp_compressed/#1/#2/2008_001274#3} &
    \includegraphics[width=\ablationw,height=\ablationh]{figures_supp_compressed/#1/#2/2008_002258#3} &
    \includegraphics[width=\ablationw,height=\ablationh]{figures_supp_compressed/#1/#2/2008_002466#3} &
    \includegraphics[width=\ablationw,height=\ablationh]{figures_supp_compressed/#1/#2/2008_002710#3} &
    \includegraphics[width=\ablationw,height=\ablationh]{figures_supp_compressed/#1/#2/2008_002943#3} &
    \includegraphics[width=\ablationw,height=\ablationh]{figures_supp_compressed/#1/#2/2008_003986#3} &
    \includegraphics[width=\ablationw,height=\ablationh]{figures_supp_compressed/#1/#2/2008_006872#3} 
    
}

\begin{figure*}[ht]
	\centering
	\footnotesize
    \begin{tabular}
	{   b{4mm}
		@{\hspace{0mm}}c@{\hspace{0.3mm}} 
		@{\hspace{0mm}}c@{\hspace{0.3mm}}
		@{\hspace{0mm}}c@{\hspace{0.3mm}} 
		@{\hspace{0mm}}c@{\hspace{0.3mm}} 
		@{\hspace{0mm}}c@{\hspace{0.3mm}} 
		@{\hspace{0mm}}c@{\hspace{0.3mm}} 
		@{\hspace{0mm}}c@{\hspace{0.3mm}} 
		@{\hspace{0mm}}c@{\hspace{0.3mm}} 
	}
	
	\vfill \rotatebox{90}{Image} \vfill &
	\ablationrow{jpegimages}{}{.jpg}\\

	\vfill \rotatebox{90}{ SCOPS (K=4)} \vfill &
	\ablationrow{pascal-car}{part_dcrf_overlay}{.jpg}\\
	
    \end{tabular}
    
    \caption{SCOPS visual results on \textbf{car} class images in the PASCAL dataset.}{}
    \label{fig:car}

\end{figure*}

\renewcommand{\ablationrow}[3]{            
	\includegraphics[width=\ablationw,height=\ablationh]{figures_supp_compressed/#1/#2/2008_003266#3} &
	\includegraphics[width=\ablationw,height=\ablationh]{figures_supp_compressed/#1/#2/2008_003373#3} &
	\includegraphics[width=\ablationw,height=\ablationh]{figures_supp_compressed/#1/#2/2008_003691#3} &
	\includegraphics[width=\ablationw,height=\ablationh]{figures_supp_compressed/#1/#2/2008_004844#3} &
	\includegraphics[width=\ablationw,height=\ablationh]{figures_supp_compressed/#1/#2/2008_004976#3} &
	\includegraphics[width=\ablationw,height=\ablationh]{figures_supp_compressed/#1/#2/2008_006748#3} &
	\includegraphics[width=\ablationw,height=\ablationh]{figures_supp_compressed/#1/#2/2008_007375#3} &
	\includegraphics[width=\ablationw,height=\ablationh]{figures_supp_compressed/#1/#2/2008_008281#3} 
	
}

\begin{figure*}[ht]
	\centering
	\footnotesize
	\begin{tabular}
		{   b{4mm}
			@{\hspace{0mm}}c@{\hspace{0.3mm}} 
			@{\hspace{0mm}}c@{\hspace{0.3mm}}
			@{\hspace{0mm}}c@{\hspace{0.3mm}} 
			@{\hspace{0mm}}c@{\hspace{0.3mm}} 
			@{\hspace{0mm}}c@{\hspace{0.3mm}} 
			@{\hspace{0mm}}c@{\hspace{0.3mm}} 
			@{\hspace{0mm}}c@{\hspace{0.3mm}} 
			@{\hspace{0mm}}c@{\hspace{0.3mm}} 
		}

		\vfill \rotatebox{90}{Image} \vfill &
		\ablationrow{jpegimages}{}{.jpg}\\

		\vfill \rotatebox{90}{ SCOPS (K=4)} \vfill &
		\ablationrow{pascal-bus}{part_dcrf_overlay}{.jpg}\\

	\end{tabular}

	\caption{SCOPS visual results on \textbf{bus} class images in the PASCAL dataset.}{}
	\label{fig:bus}
	
\end{figure*}

\renewcommand{\ablationrow}[3]{            
	\includegraphics[width=\ablationw,height=\ablationh]{figures_supp_compressed/#1/#2/2008_003703#3} &
	\includegraphics[width=\ablationw,height=\ablationh]{figures_supp_compressed/#1/#2/2008_003788#3} &
	\includegraphics[width=\ablationw,height=\ablationh]{figures_supp_compressed/#1/#2/2008_006619#3} &
	\includegraphics[width=\ablationw,height=\ablationh]{figures_supp_compressed/#1/#2/2008_006623#3} &
	\includegraphics[width=\ablationw,height=\ablationh]{figures_supp_compressed/#1/#2/2008_007145#3} &
	\includegraphics[width=\ablationw,height=\ablationh]{figures_supp_compressed/#1/#2/2008_008464#3} &
	\includegraphics[width=\ablationw,height=\ablationh]{figures_supp_compressed/#1/#2/2009_001781#3} &
	\includegraphics[width=\ablationw,height=\ablationh]{figures_supp_compressed/#1/#2/2009_003760#3} 
	
}

\begin{figure*}[ht]
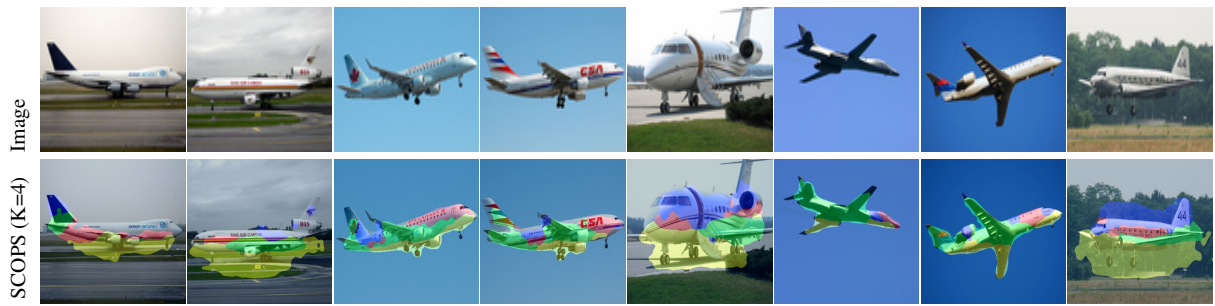

	\centering
	\footnotesize
	\begin{tabular}
		{   b{4mm}
			@{\hspace{0mm}}c@{\hspace{0.3mm}} 
			@{\hspace{0mm}}c@{\hspace{0.3mm}}
			@{\hspace{0mm}}c@{\hspace{0.3mm}} 
			@{\hspace{0mm}}c@{\hspace{0.3mm}} 
			@{\hspace{0mm}}c@{\hspace{0.3mm}} 
			@{\hspace{0mm}}c@{\hspace{0.3mm}} 
			@{\hspace{0mm}}c@{\hspace{0.3mm}} 
			@{\hspace{0mm}}c@{\hspace{0.3mm}} 
		}

		\vfill \rotatebox{90}{Image} \vfill &
		\ablationrow{jpegimages}{}{.jpg}\\

		\vfill \rotatebox{90}{ SCOPS (K=4)} \vfill &
		\ablationrow{pascal-plane}{part_dcrf_overlay}{.jpg}\\

	\end{tabular}

	\caption{SCOPS visual results on \textbf{aeroplane} class images in the PASCAL dataset.}{}
	\label{fig:plane}
	
\end{figure*}

\renewcommand{\ablationrow}[3]{            
	\includegraphics[width=\ablationw,height=\ablationh]{figures_supp_compressed/#1/#2/2008_001691#3} &
	\includegraphics[width=\ablationw,height=\ablationh]{figures_supp_compressed/#1/#2/2008_002972#3} &
	\includegraphics[width=\ablationw,height=\ablationh]{figures_supp_compressed/#1/#2/2008_003892#3} &
	\includegraphics[width=\ablationw,height=\ablationh]{figures_supp_compressed/#1/#2/2008_004822#3} &
	\includegraphics[width=\ablationw,height=\ablationh]{figures_supp_compressed/#1/#2/2008_005297#3} &
	\includegraphics[width=\ablationw,height=\ablationh]{figures_supp_compressed/#1/#2/2008_005893#3} &
	\includegraphics[width=\ablationw,height=\ablationh]{figures_supp_compressed/#1/#2/2008_007054#3} &
	\includegraphics[width=\ablationw,height=\ablationh]{figures_supp_compressed/#1/#2/2008_007075#3} 
	
}

\begin{figure*}[ht]
	\centering
	\footnotesize
	\begin{tabular}
		{   b{4mm}
			@{\hspace{0mm}}c@{\hspace{0.3mm}} 
			@{\hspace{0mm}}c@{\hspace{0.3mm}}
			@{\hspace{0mm}}c@{\hspace{0.3mm}} 
			@{\hspace{0mm}}c@{\hspace{0.3mm}} 
			@{\hspace{0mm}}c@{\hspace{0.3mm}} 
			@{\hspace{0mm}}c@{\hspace{0.3mm}} 
			@{\hspace{0mm}}c@{\hspace{0.3mm}} 
			@{\hspace{0mm}}c@{\hspace{0.3mm}} 
		}

		\vfill \rotatebox{90}{Image} \vfill &
		\ablationrow{jpegimages}{}{.jpg}\\

		\vfill \rotatebox{90}{ SCOPS (K=4)} \vfill &
		\ablationrow{pascal-motor}{part_dcrf_overlay}{.jpg}\\

	\end{tabular}

	\caption{SCOPS visual results on \textbf{motor} class images in the PASCAL dataset.}{}
	\label{fig:motor}
	
\end{figure*}

\renewcommand{\ablationrow}[3]{            
	\includegraphics[width=\ablationw,height=\ablationh]{figures_supp_compressed/#1/#2/2008_005938#3} &
	\includegraphics[width=\ablationw,height=\ablationh]{figures_supp_compressed/#1/#2/2008_006070#3} &
	\includegraphics[width=\ablationw,height=\ablationh]{figures_supp_compressed/#1/#2/2009_001172#3} &
	\includegraphics[width=\ablationw,height=\ablationh]{figures_supp_compressed/#1/#2/2009_001203#3} &
	\includegraphics[width=\ablationw,height=\ablationh]{figures_supp_compressed/#1/#2/2009_001236#3} &
	\includegraphics[width=\ablationw,height=\ablationh]{figures_supp_compressed/#1/#2/2009_004945#3} &
	\includegraphics[width=\ablationw,height=\ablationh]{figures_supp_compressed/#1/#2/2010_000189#3} &
	\includegraphics[width=\ablationw,height=\ablationh]{figures_supp_compressed/#1/#2/2010_004812#3} 
	
}

\begin{figure*}[ht]
	\centering
	\footnotesize
	\begin{tabular}
		{   b{4mm}
			@{\hspace{0mm}}c@{\hspace{0.3mm}} 
			@{\hspace{0mm}}c@{\hspace{0.3mm}}
			@{\hspace{0mm}}c@{\hspace{0.3mm}} 
			@{\hspace{0mm}}c@{\hspace{0.3mm}} 
			@{\hspace{0mm}}c@{\hspace{0.3mm}} 
			@{\hspace{0mm}}c@{\hspace{0.3mm}} 
			@{\hspace{0mm}}c@{\hspace{0.3mm}} 
			@{\hspace{0mm}}c@{\hspace{0.3mm}} 
		}

		\vfill \rotatebox{90}{Image} \vfill &
		\ablationrow{jpegimages}{}{.jpg}\\

		\vfill \rotatebox{90}{ SCOPS (K=4)} \vfill &
		\ablationrow{pascal-sheep}{part_dcrf_overlay}{.jpg}\\

	\end{tabular}

	\caption{SCOPS visual results on \textbf{sheep} class images in the PASCAL dataset.}{}
	\label{fig:sheep}
	
\end{figure*}

\renewcommand{\ablationrow}[3]{            
	\includegraphics[width=\ablationw,height=\ablationh]{figures_supp_compressed/#1/#2/2008_000552#3} &
	\includegraphics[width=\ablationw,height=\ablationh]{figures_supp_compressed/#1/#2/2008_002762#3} &
	\includegraphics[width=\ablationw,height=\ablationh]{figures_supp_compressed/#1/#2/2008_003677#3} &
	\includegraphics[width=\ablationw,height=\ablationh]{figures_supp_compressed/#1/#2/2008_004911#3} &
	\includegraphics[width=\ablationw,height=\ablationh]{figures_supp_compressed/#1/#2/2008_007004#3} &
	\includegraphics[width=\ablationw,height=\ablationh]{figures_supp_compressed/#1/#2/2008_008012#3} &
	\includegraphics[width=\ablationw,height=\ablationh]{figures_supp_compressed/#1/#2/2009_000100#3} &
	\includegraphics[width=\ablationw,height=\ablationh]{figures_supp_compressed/#1/#2/2009_003082#3} 
	
}

\begin{figure*}[ht]
	\centering
	\footnotesize
	\begin{tabular}
		{   b{4mm}
			@{\hspace{0mm}}c@{\hspace{0.3mm}} 
			@{\hspace{0mm}}c@{\hspace{0.3mm}}
			@{\hspace{0mm}}c@{\hspace{0.3mm}} 
			@{\hspace{0mm}}c@{\hspace{0.3mm}} 
			@{\hspace{0mm}}c@{\hspace{0.3mm}} 
			@{\hspace{0mm}}c@{\hspace{0.3mm}} 
			@{\hspace{0mm}}c@{\hspace{0.3mm}} 
			@{\hspace{0mm}}c@{\hspace{0.3mm}} 
		}

		\vfill \rotatebox{90}{Image} \vfill &
		\ablationrow{jpegimages}{}{.jpg}\\

		\vfill \rotatebox{90}{ SCOPS (K=3)} \vfill &
		\ablationrow{pascal-horse}{part_dcrf_overlay}{.jpg}\\

	\end{tabular}

	\caption{SCOPS visual results on \textbf{horse} class images in the PASCAL dataset.}{}
	\label{fig:horse}
	
\end{figure*}

\renewcommand{\ablationrow}[3]{            
	\includegraphics[width=\ablationw,height=\ablationh]{figures_supp_compressed/#1/#2/2008_006528#3} &
	\includegraphics[width=\ablationw,height=\ablationh]{figures_supp_compressed/#1/#2/2008_007273#3} &
	\includegraphics[width=\ablationw,height=\ablationh]{figures_supp_compressed/#1/#2/2008_008428#3} &
	\includegraphics[width=\ablationw,height=\ablationh]{figures_supp_compressed/#1/#2/2009_001145#3} &
	\includegraphics[width=\ablationw,height=\ablationh]{figures_supp_compressed/#1/#2/2009_001163#3} &
	\includegraphics[width=\ablationw,height=\ablationh]{figures_supp_compressed/#1/#2/2009_003510#3} &
	\includegraphics[width=\ablationw,height=\ablationh]{figures_supp_compressed/#1/#2/2010_002834#3} &
	\includegraphics[width=\ablationw,height=\ablationh]{figures_supp_compressed/#1/#2/2010_005258#3} 
	
}

\begin{figure*}[ht]
	\centering
	\footnotesize
	\begin{tabular}
		{   b{4mm}
			@{\hspace{0mm}}c@{\hspace{0.3mm}} 
			@{\hspace{0mm}}c@{\hspace{0.3mm}}
			@{\hspace{0mm}}c@{\hspace{0.3mm}} 
			@{\hspace{0mm}}c@{\hspace{0.3mm}} 
			@{\hspace{0mm}}c@{\hspace{0.3mm}} 
			@{\hspace{0mm}}c@{\hspace{0.3mm}} 
			@{\hspace{0mm}}c@{\hspace{0.3mm}} 
			@{\hspace{0mm}}c@{\hspace{0.3mm}} 
		}

		\vfill \rotatebox{90}{Image} \vfill &
		\ablationrow{jpegimages}{}{.jpg}\\

		\vfill \rotatebox{90}{ SCOPS (K=3)} \vfill &
		\ablationrow{pascal-cow}{part_dcrf_overlay}{.jpg}\\

	\end{tabular}

	\caption{SCOPS visual results on \textbf{cow} class images in the PASCAL dataset.}{}
	\label{fig:cow}
	
\end{figure*}

\renewcommand{\ablationrow}[4]{            
	\includegraphics[width=\ablationw,height=\ablationh]{figures_supp_compressed/#1/iter_#2/web_html/images/#3/009567#4} &
	\includegraphics[width=\ablationw,height=\ablationh]{figures_supp_compressed/#1/iter_#2/web_html/images/#3/031034#4} &
	\includegraphics[width=\ablationw,height=\ablationh]{figures_supp_compressed/#1/iter_#2/web_html/images/#3/075297#4} &
	\includegraphics[width=\ablationw,height=\ablationh]{figures_supp_compressed/#1/iter_#2/web_html/images/#3/131414#4} &
	\includegraphics[width=\ablationw,height=\ablationh]{figures_supp_compressed/#1/iter_#2/web_html/images/#3/106570#4} &
	\includegraphics[width=\ablationw,height=\ablationh]{figures_supp_compressed/#1/iter_#2/web_html/images/#3/120303#4} &
	\includegraphics[width=\ablationw,height=\ablationh]{figures_supp_compressed/#1/iter_#2/web_html/images/#3/195686#4} &
	\includegraphics[width=\ablationw,height=\ablationh]{figures_supp_compressed/#1/iter_#2/web_html/images/#3/063020#4} 
}

\begin{figure*}[ht]
	\centering
	\footnotesize
	\begin{tabular}
		{   b{4mm}
			@{\hspace{0mm}}c@{\hspace{0.3mm}} 
			@{\hspace{0mm}}c@{\hspace{0.3mm}}
			@{\hspace{0mm}}c@{\hspace{0.3mm}} 
			@{\hspace{0mm}}c@{\hspace{0.3mm}} 
			@{\hspace{0mm}}c@{\hspace{0.3mm}} 
			@{\hspace{0mm}}c@{\hspace{0.3mm}} 
			@{\hspace{0mm}}c@{\hspace{0.3mm}} 
			@{\hspace{0mm}}c@{\hspace{0.3mm}} 
		}
		\vfill \rotatebox{90}{Image} \vfill &
		\ablationrow{1111_celebawild_filter30_k8_con1e1_sep1e2_lmeqv1e2}{100000}{img}{.jpg}\\
		
		\vfill \rotatebox{90}{Saliency} \vfill &
		\ablationrow{saliency_rbd}{0}{saliency}{_rbd.jpg}\\
		
		\midrule
		
		\vfill \rotatebox{90}{ ULD~\cite{zhang2018unsupervised,thewlis2017unsupervised} } \vfill &
		\ablationrow{1111_celebawild_filter30_k8_con1e1_sep1e2_lmeqv1e2}{100000}{partooverlaydcrf}{.jpg}\\
		
		\vfill \rotatebox{90}{ DFF~\cite{collins2018deep} } \vfill &
		\ablationrow{dff-8}{0}{partooverlaydcrf}{.jpg}\\
		
		\midrule

		
		\vfill \rotatebox{90}{ SCOPS} \vfill &
		\ablationrow{1111_celebawild_filter30_dff_52_54_k8_dff1e4_con1e1_eqv1e3_lmeqv1e2}{90000}{partooverlaydcrf}{.jpg}\\
		
		\midrule
		\vfill \rotatebox{90}{ w/o $\mathcal{L}_{sc}$ } \vfill &
		\ablationrow{1111_celebawild_filter30_dff_54_k8_dff0_con1e1_eqv1e3_lmeqv1e2}{100000}{partooverlaydcrf}{.jpg}\\
		
		\vfill \rotatebox{90}{ only $\mathcal{L}_{sc}$} \vfill &
		\ablationrow{1111_celebawild_filter30_dff_54_k8_dff1e4_nosal}{100000}{partooverlaydcrf}{.jpg}\\

		\vfill \rotatebox{90}{ w/o $\mathcal{L}_{con}$ } \vfill &
		\ablationrow{1111_celebawild_filter30_dff_54_k8_dff1e4_eqv1e3_lmeqv1e2}{100000}{partooverlaydcrf}{.jpg}\\
		
		\vfill \rotatebox{90}{w/o $\mathcal{L}_{eqv}$ } \vfill &
		\ablationrow{1111_celebawild_filter30_dff_54_k8_dff1e4_con1e1}{100000}{partooverlaydcrf}{.jpg}\\
		
		\vfill \rotatebox{90}{w/o saliency} \vfill &
		\ablationrow{1111_celebawild_filter30_dff_54_k8_dff1e4_con1e1_eqv1e3_lmeqv1e2_nosal}{100000}{partooverlaydcrf}{.jpg}\\
	\end{tabular}
	
	\mycaption{Additional visual results on CelebA face images}{SCOPS produce consistent part segments compared to existing techniques. Also shown is the effect of different loss constraints.}
	\label{fig:celeba}
\end{figure*}

\renewcommand{\ablationrow}[4]{            
	\includegraphics[width=\ablationw,height=\ablationh]{figures_supp_compressed/#1/iter_#2/web_html/images/#3/189937#4} &
	\includegraphics[width=\ablationw,height=\ablationh]{figures_supp_compressed/#1/iter_#2/web_html/images/#3/197951#4} &
	\includegraphics[width=\ablationw,height=\ablationh]{figures_supp_compressed/#1/iter_#2/web_html/images/#3/134935#4} &
	\includegraphics[width=\ablationw,height=\ablationh]{figures_supp_compressed/#1/iter_#2/web_html/images/#3/177737#4} &
	\includegraphics[width=\ablationw,height=\ablationh]{figures_supp_compressed/#1/iter_#2/web_html/images/#3/102050#4} &
	\includegraphics[width=\ablationw,height=\ablationh]{figures_supp_compressed/#1/iter_#2/web_html/images/#3/080512#4} &
	\includegraphics[width=\ablationw,height=\ablationh]{figures_supp_compressed/#1/iter_#2/web_html/images/#3/147918#4} &
	\includegraphics[width=\ablationw,height=\ablationh]{figures_supp_compressed/#1/iter_#2/web_html/images/#3/176672#4} 
}

\begin{figure*}[ht]
	\centering
	\footnotesize
	\begin{tabular}
		{   b{4mm}
			@{\hspace{0mm}}c@{\hspace{0.3mm}} 
			@{\hspace{0mm}}c@{\hspace{0.3mm}}
			@{\hspace{0mm}}c@{\hspace{0.3mm}} 
			@{\hspace{0mm}}c@{\hspace{0.3mm}} 
			@{\hspace{0mm}}c@{\hspace{0.3mm}} 
			@{\hspace{0mm}}c@{\hspace{0.3mm}} 
			@{\hspace{0mm}}c@{\hspace{0.3mm}} 
			@{\hspace{0mm}}c@{\hspace{0.3mm}} 
		}
		\vfill \rotatebox{90}{Image} \vfill &
		\ablationrow{1111_celebawild_filter30_k8_con1e1_sep1e2_lmeqv1e2}{100000}{img}{.jpg}\\
		
		\vfill \rotatebox{90}{Saliency} \vfill &
		\ablationrow{saliency_rbd}{0}{saliency}{_rbd.jpg}\\
		
		\midrule
		
		\vfill \rotatebox{90}{ ULD~\cite{zhang2018unsupervised,thewlis2017unsupervised} } \vfill &
		\ablationrow{1111_celebawild_filter30_k8_con1e1_sep1e2_lmeqv1e2}{100000}{partooverlaydcrf}{.jpg}\\
		
		\vfill \rotatebox{90}{ DFF~\cite{collins2018deep} } \vfill &
		\ablationrow{dff-8}{0}{partooverlaydcrf}{.jpg}\\
		
		\midrule

		
		\vfill \rotatebox{90}{ SCOPS} \vfill &
		\ablationrow{1111_celebawild_filter30_dff_52_54_k8_dff1e4_con1e1_eqv1e3_lmeqv1e2}{90000}{partooverlaydcrf}{.jpg}\\
		
		\midrule
		\vfill \rotatebox{90}{ w/o $\mathcal{L}_{sc}$ } \vfill &
		\ablationrow{1111_celebawild_filter30_dff_54_k8_dff0_con1e1_eqv1e3_lmeqv1e2}{100000}{partooverlaydcrf}{.jpg}\\
		
		\vfill \rotatebox{90}{ only $\mathcal{L}_{sc}$} \vfill &
		\ablationrow{1111_celebawild_filter30_dff_54_k8_dff1e4_nosal}{100000}{partooverlaydcrf}{.jpg}\\

		\vfill \rotatebox{90}{ w/o $\mathcal{L}_{con}$ } \vfill &
		\ablationrow{1111_celebawild_filter30_dff_54_k8_dff1e4_eqv1e3_lmeqv1e2}{100000}{partooverlaydcrf}{.jpg}\\
		
		\vfill \rotatebox{90}{w/o $\mathcal{L}_{eqv}$ } \vfill &
		\ablationrow{1111_celebawild_filter30_dff_54_k8_dff1e4_con1e1}{100000}{partooverlaydcrf}{.jpg}\\
		
		\vfill \rotatebox{90}{w/o saliency} \vfill &
		\ablationrow{1111_celebawild_filter30_dff_54_k8_dff1e4_con1e1_eqv1e3_lmeqv1e2_nosal}{100000}{partooverlaydcrf}{.jpg}\\
	\end{tabular}
	
	\mycaption{Additional visual results on CelebA face images}{SCOPS produce consistent part segments compared to existing techniques. Also shown is the effect of different loss constraints.}
	\label{fig:celeba2}
\end{figure*}

\renewcommand{\cubrow}[2]{ 
	
	\includegraphics[width=\ablationw,height=\ablationh]{figures_supp_compressed/cub/002/#1/laysan_albatross_0029_482.#2} &
	\includegraphics[width=\ablationw,height=\ablationh]{figures_supp_compressed/cub/002/#1/laysan_albatross_0033_658.#2} &
	\includegraphics[width=\ablationw,height=\ablationh]{figures_supp_compressed/cub/002/#1/laysan_albatross_0034_628.#2} &
	\includegraphics[width=\ablationw,height=\ablationh]{figures_supp_compressed/cub/002/#1/laysan_albatross_0037_699.#2} &
	\includegraphics[width=\ablationw,height=\ablationh]{figures_supp_compressed/cub/002/#1/laysan_albatross_0044_784.#2} &
	\includegraphics[width=\ablationw,height=\ablationh]{figures_supp_compressed/cub/002/#1/laysan_albatross_0058_637.#2} &
	\includegraphics[width=\ablationw,height=\ablationh]{figures_supp_compressed/cub/002/#1/laysan_albatross_0061_563.#2} &
	\includegraphics[width=\ablationw,height=\ablationh]{figures_supp_compressed/cub/002/#1/laysan_albatross_0085_564.#2} 
}

\begin{figure*}[ht]
	\centering
	\footnotesize
	\begin{tabular}
		{   b{4mm}
			@{\hspace{0mm}}c@{\hspace{0.3mm}} 
			@{\hspace{0mm}}c@{\hspace{0.3mm}}
			@{\hspace{0mm}}c@{\hspace{0.3mm}} 
			@{\hspace{0mm}}c@{\hspace{0.3mm}} 
			@{\hspace{0mm}}c@{\hspace{0.3mm}} 
			@{\hspace{0mm}}c@{\hspace{0.3mm}} 
			@{\hspace{0mm}}c@{\hspace{0.3mm}} 
			@{\hspace{0mm}}c@{\hspace{0.3mm}} 
		}

		\vfill \rotatebox{90}{ Image } \vfill &
		\cubrow{images}{jpg} \\
		\vfill \rotatebox{90}{ ULD~\cite{zhang2018unsupervised,thewlis2017unsupervised}} \vfill &
		\cubrow{uld}{jpg} \\
		\vfill \rotatebox{90}{ DFF~\cite{collins2018deep} } \vfill &
		\cubrow{dff}{jpg} \\
		\vfill \rotatebox{90}{ SCOPS } \vfill &
		\cubrow{scops}{jpg} \\

	\end{tabular}

	\mycaption{Additional visual results on CUB bird images}{SCOPS is robust to pose and camera variations while having better boundary adherence compared to other techniques.}
	\label{fig:cub}
	
\end{figure*}

\renewcommand{\cubrow}[2]{            
	
	\includegraphics[width=\ablationw,height=\ablationh]{figures_supp_compressed/cub/002/#1/laysan_albatross_0091_602.#2} &
	\includegraphics[width=\ablationw,height=\ablationh]{figures_supp_compressed/cub/002/#1/laysan_albatross_0093_725.#2} &
	\includegraphics[width=\ablationw,height=\ablationh]{figures_supp_compressed/cub/002/#1/laysan_albatross_0094_1013.#2} &
	\includegraphics[width=\ablationw,height=\ablationh]{figures_supp_compressed/cub/002/#1/laysan_albatross_0096_673.#2} &
	\includegraphics[width=\ablationw,height=\ablationh]{figures_supp_compressed/cub/002/#1/laysan_albatross_0099_869.#2} &
	\includegraphics[width=\ablationw,height=\ablationh]{figures_supp_compressed/cub/002/#1/laysan_albatross_0100_735.#2} &
	\includegraphics[width=\ablationw,height=\ablationh]{figures_supp_compressed/cub/002/#1/laysan_albatross_0102_611.#2} &
	\includegraphics[width=\ablationw,height=\ablationh]{figures_supp_compressed/cub/002/#1/laysan_albatross_0103_504.#2} 
}

\begin{figure*}[ht]
	\centering
	\footnotesize
	\begin{tabular}
		{   b{4mm}
			@{\hspace{0mm}}c@{\hspace{0.3mm}} 
			@{\hspace{0mm}}c@{\hspace{0.3mm}}
			@{\hspace{0mm}}c@{\hspace{0.3mm}} 
			@{\hspace{0mm}}c@{\hspace{0.3mm}} 
			@{\hspace{0mm}}c@{\hspace{0.3mm}} 
			@{\hspace{0mm}}c@{\hspace{0.3mm}} 
			@{\hspace{0mm}}c@{\hspace{0.3mm}} 
			@{\hspace{0mm}}c@{\hspace{0.3mm}} 
		}

		\vfill \rotatebox{90}{ Image } \vfill &
		\cubrow{images}{jpg} \\
		\vfill \rotatebox{90}{ ULD~\cite{zhang2018unsupervised,thewlis2017unsupervised} } \vfill &
		\cubrow{uld}{jpg} \\
		\vfill \rotatebox{90}{ DFF~\cite{collins2018deep} } \vfill &
		\cubrow{dff}{jpg} \\
		\vfill \rotatebox{90}{ SCOPS } \vfill &
		\cubrow{scops}{jpg} \\

	\end{tabular}

	\mycaption{Additional visual results on CUB bird images}{SCOPS is robust to pose and camera variations while having better boundary adherence compared to other techniques.}
	\label{fig:cub2}
	
\end{figure*}

\end{document}